\documentclass[preprint,journal]{IEEEtai}

\usepackage[colorlinks,urlcolor=blue,linkcolor=blue,citecolor=blue]{hyperref}

\usepackage{color,array}

\usepackage{graphicx}

\usepackage{multirow}
\usepackage{booktabs} % For formal tables
\usepackage{array}    % For table wrapping
\usepackage{amsmath}
\usepackage{xcolor}         % colors 
\usepackage{amsfonts}
\usepackage{bm}
\usepackage{tikz}
\usetikzlibrary{arrows.meta}
\usepackage{cleveref}
\usepackage[numbers]{natbib}
\usepackage{subcaption}
\usepackage{graphicx}
\let\labelindent\relax
\usepackage{enumitem}

\usepackage{algorithm}
\usepackage{algpseudocode}
\usetikzlibrary{positioning, shapes.geometric, calc, decorations.pathreplacing}
\newcommand{\context}[1]{} 

\newcommand{\algo}{\textsc{sachi}}

\begin{document}

\title{\algo: \underline{S}tructured \underline{A}gent \underline{C}oordination via \underline{H}olistic \underline{I}nformation Integration in Multi-Agent Reinforcement Learning} 

\author{
\IEEEauthorblockN{
Nikunj Gupta\textsuperscript{1},
James Zachary Hare\textsuperscript{2},
Jesse Milzman\textsuperscript{2},
Rajgopal Kannan\textsuperscript{3},
Viktor Prasanna\textsuperscript{1}\\
}
\IEEEauthorblockA{
\textsuperscript{1}University of Southern California, Los Angeles, CA, USA \\
\textsuperscript{2}DEVCOM Army Research Laboratory, Adelphi, MD, USA \\
\textsuperscript{3}DEVCOM Army Research Office, Research Triangle Park, North Carolina, USA \\
\textit{Emails: \{nikunj, prasanna\}@usc.edu \{james.z.hare, jesse.m.milzman, rajgopal.kannan\}.civ@army.mil} 
}
}

% \markboth{Journal of IEEE Transactions on Artificial Intelligence, Vol. 00, No. 0, Month 2020}
% {First A. Author \MakeLowercase{\textit{et al.}}: Bare Demo of IEEEtai.cls for IEEE Journals of IEEE Transactions on Artificial Intelligence}

\maketitle

\begin{abstract}
Cooperative multi-agent reinforcement learning agents that act on partial local observations face a fundamental information bottleneck: the knowledge needed to select jointly optimal actions is scattered across the team, yet each agent must commit to a decision without access to its teammates' observations, intentions, or chosen actions. Existing methods either ignore this bottleneck, compress it into a scalar mixing signal, or route around it with learned communication channels. Framing action coordination as a problem of structured information integration among agents, we propose \textit{structured agent coordination via holistic information integration}, or \algo, in which graph transformer convolutions over an inter-agent coordination graph enrich each agent's representation with receiver-sensitive, content-dependent signals from teammates prior to action selection. We evaluate \algo\ across five cooperative tasks spanning spatial, communicative, and adversarial coordination challenges against twelve baselines. \algo\ consistently matches or outperforms the best baseline on every task, and rigorous aggregate statistical analyses, including normalized metrics with bootstrap confidence intervals, Friedman ranking, and performance profiling, confirm that this advantage is statistically significant, robust across environments, and not attributable to increased model capacity. Parameter-matched ablations further trace the source of the gains to a single architectural property: the degree of content-dependence in the message-passing operator.

\end{abstract}

\begin{IEEEImpStatement}
This paper advances the scientific understanding of coordination in multi-agent systems by framing action coordination as a problem of structured information integration. The methods developed here are intended to improve the performance of cooperative autonomous systems in settings where effective teamwork is essential, including robotics, logistics, and resource allocation. We foresee no immediate ethical concerns arising from this work. As with any advance in autonomous decision-making, the broader societal implications depend largely on the contexts in which these methods are deployed, and we encourage responsible application. 
\end{IEEEImpStatement}

\begin{IEEEkeywords}
Multi-agent reinforcement learning, cooperative agents, action coordination, coordination graphs, graph neural networks, graph transformers.
\end{IEEEkeywords}

\section{Introduction}
\label{sec:Introduction} 
\context{introducing coordination, cooperative MARL, action coordination in MARL, mentioning influx of algorithms and applications}
Coordination, the alignment of individual decisions toward a shared objective, is among the oldest and most consequential problems in the study of organized systems~\cite{mooney1947principles}. From assembly lines to air-traffic control, success depends not on any single actor in isolation but on the degree to which many actors select mutually compatible actions under uncertainty. Multi-agent reinforcement learning (MARL) inherits this challenge in full generality: a team of agents must learn a joint policy under which each agent's action is individually executable yet collectively coherent, often from partial and local observations of a shared environment~\cite{oliehoek2016concise, bernstein2002complexity}. The past decade has seen a remarkable influx of cooperative MARL algorithms, spanning value decomposition~\cite{sunehag2017value, rashid2020monotonic}, communication learning, and policy-gradient methods~\cite{lowe2017multi, yu2022surprising}, motivated by applications in autonomous driving, warehouse logistics, robotic manipulation, and network routing~\cite{li2022applications, oroojlooy2023review, wang2022distributed}. Yet across this landscape, a persistent difficulty remains: how should a joint policy be structured so that each agent's locally-computed action is compatible with the actions of teammates it cannot observe?

\context{illustrating coordination problems with examples, from simple to complex}
To appreciate why action coordination is difficult, it helps to first consider a setting where it is not. Imagine a team of delivery drones operating in non-overlapping zones of a city: each drone plans its own route independently, and the fleet's total throughput is simply the sum of individual deliveries. A policy that maximizes each agent's individual return also maximizes the team's, so no coordination of actions is required. Now extend the scenario. Suppose zones overlap and two drones approaching the same rooftop must yield to avoid collision; the action each drone should take now depends on the action the other is taking, and a policy that ignores this coupling will produce collisions no matter how well each agent optimizes locally~\cite{marek2025collision}. Add heterogeneity: heavy packages demand larger drones and narrow alleys require smaller ones, so the optimal action for one agent is contingent on the assignments of others. Introduce time coupling: a drone occupying a rooftop landing pad blocks a second drone carrying a perishable item, so even the order in which agents act must be jointly determined. At each extension, individual action optimality becomes insufficient. What matters is whether the joint action profile, the vector of all agents' simultaneous choices, is coherent as a whole. This joint coherence under decentralized execution is the defining challenge of action coordination in cooperative MARL. 

\context{the key impediment: team decisions where all agents share the same preferences over outcomes (no conflict of interest), yet reaching the best outcome is hard}
One of the most subtle impediments to action coordination arises in settings where agents are fully aligned: every member of the team shares the same preference ranking over group outcomes, and any joint action profile that raises one agent's payoff raises everyone's. In such cooperative teams, the difficulty is not conflicting incentives but the structural problem of producing a coherent joint action from decentralized decisions~\cite{oliehoek2016concise}. Each agent controls only its own action and observes only a local view of the environment, yet the value of that action depends on the actions simultaneously chosen by its teammates. A search-and-rescue squad may unanimously want to cover a disaster site as quickly as possible, yet if each rescuer independently selects the nearest uncovered sector, the resulting joint action profile leaves critical zones unvisited regardless of how rational each individual choice appears in isolation. Producing a jointly optimal action profile, therefore, requires each agent's policy to be conditioned not only on its own observations but also on information that makes teammates' actions predictable and mutually consistent. The central challenge, then, is how to act when the information needed to select a compatible action is distributed across agents with no mechanism to assemble it.

\begin{figure}
\centerline{\includegraphics[width=\linewidth]{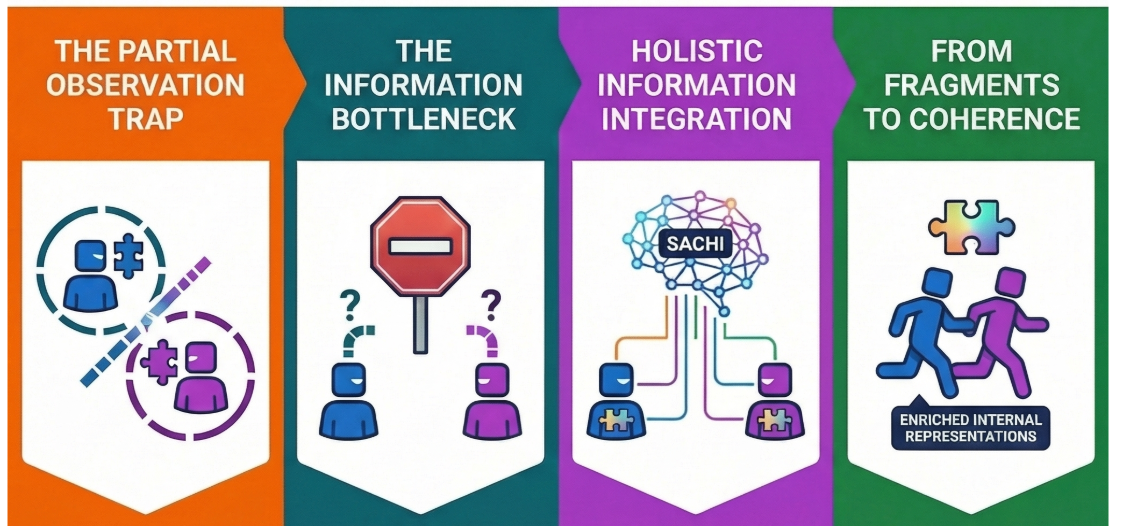}} 
\caption{\textbf{Solving the MARL Information Bottleneck.} (a) Each agent holds only a fragment of the information needed to determine the globally optimal joint action; (b) Since agents cannot see their teammates' observations or intentions, their independent ``rational" choices often lead to group failure; (c) \algo\ resolves this by allowing agents to intelligently filter and ``borrow" relevant context from teammates through graph transformer layers; and (d) Enriched internal representations allow agents to act independently at runtime while maintaining perfect joint alignment.}
% \vspace{-15pt}
\end{figure} 

\context{In this paper, we propose SACHI: focusing on enhanced agent information integration via graph transformers over coordination/agent interaction graphs} 
In this paper, we address this action-coordination challenge directly by targeting the information bottleneck that prevents decentralized policies from producing coherent joint actions. We propose \textbf{S}tructured \textbf{A}gent \textbf{C}oordination via \textbf{H}olistic \textbf{I}nformation \textbf{I}ntegration (\algo), where \emph{holistic} reflects the principle that effective coordination does not require an agent to collect all available information from all teammates indiscriminately, but rather to intelligently accumulate precisely the coordination-relevant signals it needs from its neighbors. It must filter, weight, and integrate incoming information in a manner that is sensitive to the agent's own state and role. \algo\ achieves this by enriching each agent's policy input through graph transformer convolutions~\cite{shi2020masked} over an inter-agent coordination graph, in which content-dependent attention determines not only which neighboring agents contribute to a given representation but how much and in what form their information is incorporated. The coordination graph encodes \emph{which} agents' actions are coupled; the graph transformer determines \emph{what} information flows between them and \emph{how} each agent integrates incoming signals given its own state. The result is a set of enriched representations (one per agent) that carry precisely the information needed for each agent's locally computed action to be compatible with its teammates', enabling decentralized policies to jointly solve the coordination problem. By resolving action coordination at the representation level, \algo\ provides a principled approach to producing coherent joint behavior from individually acting agents. 

% Paragraph 5: contributions
Our contributions are as follows. First, we offer a conceptual reframing of action coordination in cooperative MARL as a policy-level problem of structured information integration, arguing that decentralized agents cannot reliably produce coherent joint actions unless each agent's representation reflects the broader joint context. Second, we propose \algo, a practical instantiation of this perspective that enriches agent representations through graph transformer message passing over a coordination graph, and combines these representations with value decomposition to support fully decentralized execution at test time. Third, we provide an empirical evaluation across five cooperative multi-agent tasks spanning spatial, communicative, and adversarial coordination challenges, showing consistent improvements over established baselines and confirming through ablation studies that the gains are attributable to the coordination mechanism itself rather than increased model capacity.

\section{Related Works}
\label{sec:related_works}

\subsection{Cooperative MARL} Cooperative multi-agent reinforcement learning addresses 
the problem of training a team of agents to maximize a shared objective through 
interaction with a common environment~\citep{busoniu2008comprehensive, 
hernandez2019survey}. Early approaches extended single-agent RL directly to the 
multi-agent setting, either by training agents independently~\citep{tan1993multi, 
matignon2012independent} or by treating the joint action space as a single agent's 
action space~\citep{claus1998dynamics}, both of which scale poorly as the number of 
agents grows. A central difficulty in cooperative MARL is the non-stationarity 
introduced by simultaneously learning agents: from any single agent's perspective, 
the environment appears to change as its teammates update their policies, violating 
the stationarity assumptions underlying standard RL convergence 
guarantees~\citep{hernandez2019survey}. More recent work has addressed this through 
communication learning~\citep{foerster2016learning, sukhbaatar2016learning, 
das2019tarmac}, role-based decomposition~\citep{wang2020roma, li2021celebrating}, 
and emergent coordination strategies~\citep{lowe2017multi, yu2022surprising}. A 
growing body of work also examines scalability and generalization in large 
teams~\citep{rashid2020monotonic, de2020independent}, as well as the theoretical 
foundations of multi-agent credit assignment~\citep{kuba2022trust}. The present work 
situates itself within this landscape by targeting action coordination as a 
first-class problem, treating it as a policy-level challenge that requires structured 
information integration among agents rather than a byproduct of reward maximization. 

\subsection{Centralized Training with Decentralized Execution} The centralized 
training with decentralized execution (CTDE) paradigm~\citep{oliehoek2016concise, 
kraemer2016multi} has become the dominant framework for cooperative MARL, allowing 
agents to exploit global information during training while retaining the ability to 
act independently at execution time. Within CTDE, value decomposition methods address 
the credit assignment problem by factoring the joint action-value function into 
per-agent components. VDN~\citep{sunehag2017value} assumes the joint value is an 
additive sum of individual values, while QMIX~\citep{rashid2020monotonic} relaxes 
this to a monotonic combination conditioned on the global state, enabling richer 
joint value representations while preserving the consistency between centralized 
training and decentralized execution. QTRAN~\citep{son2019qtran} and 
QPLEX~\citep{wang2020qplex} further extend the representable class of joint value 
functions at the cost of increased complexity. On the actor-critic side, 
COMA~\citep{foerster2018counterfactual} uses a centralized critic with a 
counterfactual baseline for credit assignment, and MAPPO~\citep{yu2022surprising} 
demonstrates that a shared centralized critic with proximal policy optimization can 
achieve strong performance across a wide range of cooperative tasks. The present work 
builds on CTDE and adopts QMIX as its value decomposition backbone, augmenting the 
agent representation stage with structured inter-agent information exchange prior to 
action-value estimation. 

\subsection{Graph-Based Multi-Agent Reinforcement Learning} Graph-based methods have 
attracted considerable interest in MARL as a natural framework for modeling structured 
agent interactions. Coordination graphs (CGs)~\citep{guestrin2002coordinated} 
represent the factored structure of the joint value function as a graph over agents, 
enabling exact or approximate inference of the optimal joint action via message 
passing. Deep coordination graphs (DCG)~\citep{bohmer2020deep} integrate CGs with 
deep RL by learning per-edge payoff functions and performing max-sum inference at 
execution time. DICG~\citep{li2020deepimplicit} introduces a dynamic interaction 
graph inferred from agent observations, allowing the coordination topology to adapt 
to the current state. LTSCG~\citep{yang2022self} further learns a sparse coordination 
graph through a self-organizing mechanism, reducing unnecessary coordination overhead, 
and SOPCG~\citep{wang2022contextaware} conditions the coordination graph on the 
global context to produce task-relevant interaction structures. More recent work has explored richer forms of relational reasoning and adaptive 
coordination. DMCG~\citep{gupta2025deep} models higher-order and indirect agent 
dependencies through deep meta coordination graphs, enabling coordination beyond 
pairwise interactions. HAMMER~\citep{gupta2025hammer} incorporates hierarchical 
and multi-level relational abstractions to improve scalable coordination in 
complex multi-agent environments. TIGER~\citep{gupta2025tiger} studies temporal 
and interaction-aware graph representations for capturing evolving coordination 
patterns over time, while AGP~\citep{gupta2026action} introduces action-level 
graph priors that explicitly model dependencies between agent-action pairs to 
improve fine-grained coordination. Beyond coordination 
graphs, graph neural networks have been applied more broadly in MARL to model agent 
interactions~\citep{Jiang2020Graph, malysheva2018deep}, encode relational inductive 
biases~\citep{battaglia2018relational}, and support scalable multi-agent 
communication~\citep{nayak2023informarl}. Graph attention 
networks~\citep{veličković2018graph} and graph transformers~\citep{shi2020masked} 
have been explored in related settings for their ability to perform content-dependent 
aggregation over graph neighborhoods. The present work draws on this literature but 
departs from prior CG methods by replacing standard graph convolutions with graph 
transformer convolutions, enabling receiver-dependent, content-adaptive message 
passing that directly addresses the information-integration requirements of action 
coordination.

\section{Preliminaries}
\label{sec:preliminaries}

\subsection{The Cooperative Multi-Agent Model} We formalize the cooperative multi-agent 
setting as a Decentralized Partially Observable Markov Decision Process 
(Dec-POMDP)~\citep{oliehoek2016concise}, defined by the tuple $\langle \mathcal{S}, 
\mathcal{N}, \mathcal{A}, \mathcal{O}, P, R, \gamma \rangle$. 
Here $\mathcal{S}$ is the set of global environment states, $\mathcal{N} = \{1, \dots, n\}$ 
is the finite set of agents, and $\mathcal{A} = \mathcal{A}_1 \times \cdots \times 
\mathcal{A}_n$ is the joint action space, where $\mathcal{A}_i$ is the local action space 
of agent $i$. At each timestep $t$, the environment occupies a global state $s^t \in 
\mathcal{S}$ that is not directly accessible to any agent. Instead, each agent $i$ receives 
a private local observation $o_i^t \in \mathcal{O}$ drawn from an observation function 
that maps the global state to each agent's partial view of the environment. Each agent 
selects an action $a_i^t \in \mathcal{A}_i$, forming a joint action $\mathbf{a}^t = 
(a_1^t, \dots, a_n^t) \in \mathcal{A}$. The environment transitions according to 
$P(s^{t+1} \mid s^t, \mathbf{a}^t)$, and all agents receive a shared team reward 
$r^t = R(s^t, \mathbf{a}^t)$. Because each agent observes only $o_i^t$ rather than the 
global state, policies are conditioned on each agent's action-observation history 
$\tau_i^t = (o_i^0, a_i^0, \dots, o_i^t)$, which summarizes all locally available 
information up to timestep $t$. The individual policy $\pi_i(a_i \mid \tau_i)$ maps this 
history to a distribution over local actions, and the joint policy $\boldsymbol{\pi} = 
(\pi_1, \dots, \pi_n)$ induces a distribution over joint action profiles at each timestep. 
The team objective is to find a joint policy that maximizes the expected discounted return:
\begin{equation}
    J(\boldsymbol{\pi}) = \mathbb{E}\!\left[\sum_{t=0}^{T} \gamma^t r^t\right],
    \label{eq:objective}
\end{equation}
where $\gamma \in [0,1)$ is the discount factor. The central difficulty is that, although 
the reward depends on the joint action, each agent must select its action based solely on 
its own history, with no direct access to the observations, histories, or intended actions 
of its teammates.

\subsection{Value Decomposition under CTDE}
Within the CTDE paradigm~\citep{oliehoek2016concise, kraemer2016multi}, value decomposition 
methods factorize the joint action-value function $Q_{\text{tot}}(\boldsymbol{\tau}, 
\mathbf{a}, s)$ into per-agent components to enable tractable decentralized action 
selection. QMIX~\citep{rashid2020monotonic} parameterizes $Q_{\text{tot}}$ as a monotonic 
combination of individual action-value functions $Q_i(\tau_i, a_i)$:
\begin{equation}
    Q_{\text{tot}}(\boldsymbol{\tau}, \mathbf{a}, s) = f_{\text{mix}}\!\left(Q_1(\tau_1, 
    a_1), \dots, Q_n(\tau_n, a_n);\, s\right),
    \label{eq:qmix}
\end{equation}
where $f_{\text{mix}}$ is a feedforward network whose non-negative weights are generated 
by hypernetworks conditioned on the global state $s$, enforcing the monotonicity constraint 
$\partial Q_{\text{tot}} / \partial Q_i \geq 0$ for all $i$. This constraint guarantees 
that the joint greedy action $\arg\max_{\mathbf{a}} Q_{\text{tot}}$ decomposes into 
independent per-agent greedy actions $\arg\max_{a_i} Q_i$, maintaining consistency between 
centralized training and decentralized execution. The system is trained end-to-end by 
minimizing a TD loss with double Q-learning~\citep{van2016deep} against a periodically 
updated target network. In \algo, QMIX serves as the value decomposition backbone, 
receiving per-agent $Q_i$ values computed from coordination-enriched representations 
constructed by the graph transformer layers described in the next section.

\subsection{From Graph Convolutions to Graph Transformers}
Graph neural networks (GNNs) provide a general mechanism for propagating information 
between agents by treating the agent set as nodes and inter-agent relationships as edges 
of a graph $\mathcal{G} = (\mathcal{N}, \mathcal{E})$. Each agent $i$ is associated with a learned representation (embedding) 
$\mathbf{h}_i^{(\ell)} \in \mathbb{R}^d$ at layer $\ell$, initialized 
as $\mathbf{h}_i^{(0)} = \phi_{\text{enc}}(\tau_i)$ from its local 
action-observation history and iteratively refined through message passing. At each layer $\ell$, a GNN 
updates each agent's representation by aggregating messages from its neighbors:
\begin{equation}
    \mathbf{h}_i^{(\ell+1)} = \phi\!\left(\mathbf{h}_i^{(\ell)},\; 
    \bigoplus_{j \in \mathcal{N}(i)} \psi\!\left(\mathbf{h}_i^{(\ell)}, 
    \mathbf{h}_j^{(\ell)}\right)\right),
    \label{eq:gnn}
\end{equation}
where $\phi$ and $\psi$ are learnable functions, $\mathcal{N}(i)$ is the neighbor set of 
agent $i$, and $\bigoplus$ is a permutation-invariant aggregation operator. Standard graph 
convolutions~\citep{kipf2017semi} instantiate $\psi$ as a fixed linear transformation 
applied uniformly to all neighbors, which treats every incoming message as equally 
informative regardless of the receiver's current state. Graph attention 
networks~\citep{veličković2018graph} improve on this by computing a scalar attention 
weight $\alpha_{ij}$ for each neighbor, but the weight depends only on the sender's 
features and does not adapt to the receiver's query. Graph transformer 
convolutions~\citep{shi2020masked} address both limitations by computing attention via 
scaled dot-product between a query derived from the receiver and a key derived from the 
sender:
\begin{equation}
    \alpha_{ij} = \frac{\exp\!\left((\mathbf{W}_Q\, \mathbf{h}_i)^\top (\mathbf{W}_K\, 
    \mathbf{h}_j) / \sqrt{d}\right)}{\sum_{k \in \mathcal{N}(i)} \exp\!\left((\mathbf{W}_Q 
    \,\mathbf{h}_i)^\top (\mathbf{W}_K\, \mathbf{h}_k) / \sqrt{d}\right)},
    \label{eq:gt_attention}
\end{equation}
where $\mathbf{W}_Q$ and $\mathbf{W}_K$ are learnable query and key projections and $d$ 
is the embedding dimension. This formulation makes the attention weight $\alpha_{ij}$ 
dependent on both the sender's state and the receiver's current needs, producing 
content-dependent, asymmetric message passing in which the same neighbor may contribute 
different information to different receivers. As we argue in the next section, this 
property directly addresses the three core sub-problems of action coordination: which 
agents should communicate, what information they should send, and how each receiver should 
integrate incoming signals given its own state.

\section{Method: \algo}
\label{sec:methodology}

The preceding discussion identified coordination failure in cooperative teams as a failure
of information integration: each agent holds a partial view of the world, and no mechanism
exists to assemble those views into representations that support jointly coherent action
selection. \algo\ addresses this directly. The central idea is to enrich each agent's
internal representation with coordination-relevant information from its teammates
before action selection, so that standard per-agent Q-value maximization produces
coordinated behavior without requiring explicit run-time communication. We accomplish this
through a pipeline of four stages: input construction, observation encoding, graph
transformer message passing over an inter-agent coordination graph, and per-agent
action-value estimation with centralized value mixing, described below and summarized in
Algorithm~\ref{alg:sachi}. 

\begin{figure*}
\centerline{\includegraphics[width=\linewidth]{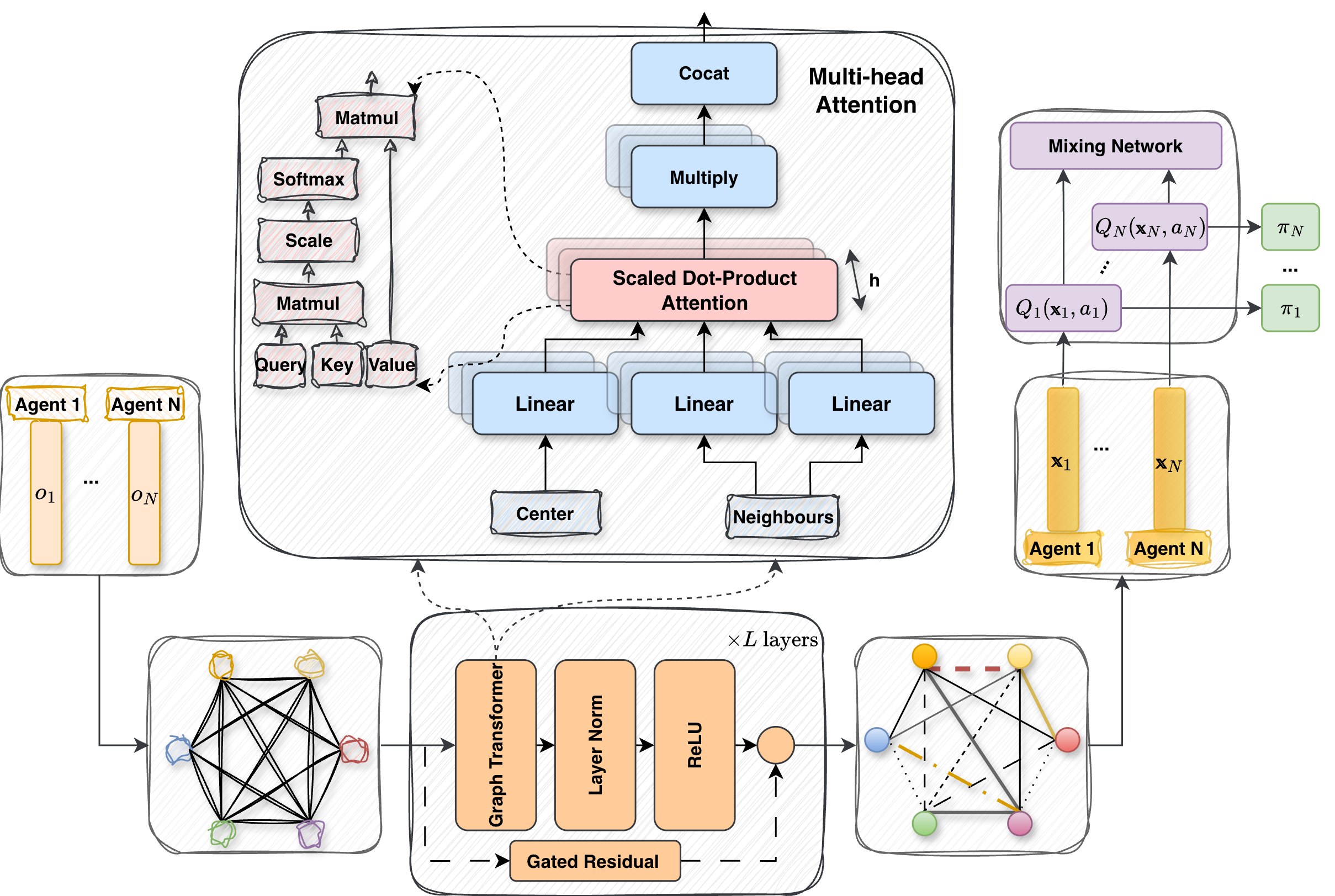}}
\caption{\textbf{Overview of \algo}. Local observations are encoded into agent embeddings, refined 
through soft-attention-modulated graph transformer message passing, and mapped to 
per-agent Q-values. The model performs receiver-dependent information integration 
prior to action selection, allowing coordinated behavior to emerge under decentralized 
execution.}
\end{figure*}

\paragraph{Input construction}
At each timestep $t$, agent $i$ receives a local observation $o_i^t$ from the environment.
We augment this with a one-hot agent identifier $\mathbf{e}_i \in \{0,1\}^n$ and form the
input:
\begin{equation}
    \mathbf{x}_i^t = \left[o_i^t \;\|\; \mathbf{e}_i\right] \in \mathbb{R}^{d_{\text{in}}},
    \label{eq:input}
\end{equation}
where $\|$ denotes concatenation. The identity vector enables the shared network to learn
agent-specific behavior or a form of implicit role specialization, while retaining full
parameter sharing across the team.

\paragraph{From observations to embeddings}
Each agent's input is mapped to a latent embedding through a shared encoder
$\phi_{\text{enc}}$, implemented as a two-layer MLP with hidden dimension $d_h = 128$:
\begin{equation}
    \mathbf{h}_i^{(0)} = \phi_{\text{enc}}(\mathbf{x}_i^t) \in \mathbb{R}^{d},
    \label{eq:encoder}
\end{equation}
where $d = d_{\text{in}}$ so that the embedding dimension is commensurate with the input
space. At this stage, each embedding reflects only agent $i$'s local information; the
next stage is designed to break this isolation.

subsequent \paragraph{Graph topology and soft-attention modulation}
We define a coordination graph $\mathcal{G} = (\mathcal{N}, \mathcal{E})$ over the agent
set, represented as a dense adjacency matrix $\mathbf{A} \in \{0,1\}^{n \times n}$. We
use a fully connected graph $\mathbf{A} = \mathbf{1}\mathbf{1}^\top$, so that every agent
pair has the opportunity to exchange information. Rather than engineering or learning the
topology explicitly, we apply a soft-attention modulation layer to the encoded embeddings,
producing a dense attention matrix $\mathbf{G} \in \mathbb{R}^{n \times n}$ that
reweights the coordination graph:
\begin{equation}
    \mathbf{G} = \mathrm{softmax}\!\left(
        \phi_{\text{att}}(\mathbf{H}^{(0)})\; (\mathbf{H}^{(0)})^\top
    \right),
    \label{eq:attention_modulation}
\end{equation}
where $\mathbf{H}^{(0)} \in \mathbb{R}^{n \times d}$ is the matrix of stacked initial
embeddings and $\phi_{\text{att}}$ is a learned linear projection. The effective adjacency
used for message passing is then $\tilde{\mathbf{A}} = \mathbf{A} \odot \mathbf{G}$, where
$\odot$ denotes element-wise multiplication. This mechanism allows \algo\ to dynamically
suppress irrelevant edges and amplify informative ones before graph transformer message
passing begins. This form of attention-based graph reweighting, commonly used in prior graph-based MARL \cite{li2020deepimplicit,naderializadeh2020graph,malysheva2018deep},
provides a coarse, global estimate of agent relevance by modulating which interactions are
emphasized. In \algo, this serves as an initial filtering stage, after which the subsequent
graph transformer layers perform fine-grained, receiver-dependent, content-conditioned
message passing. 
% providing a coarse, global view of agent relevance that complements the fine-grained, content-dependent attention computed within each graph transformer layer.

\paragraph{Graph transformer message passing}
The core of \algo\ is a stack of $L$ graph transformer convolution layers~\citep{
shi2020masked} that iteratively refine agent embeddings by passing messages along the edges
of $\tilde{\mathbf{A}}$. At layer $\ell$, the embedding of agent $i$ is updated as:
\begin{equation}
    \mathbf{h}_i^{(\ell+1)} = \tanh\!\left(
        \mathbf{W}_1^{(\ell)}\, \mathbf{h}_i^{(\ell)}
        \;+\;
        \sum_{j \in \mathcal{N}(i)} \alpha_{ij}^{(\ell)}\;
        \mathbf{W}_2^{(\ell)}\, \mathbf{h}_j^{(\ell)}
        \;+\; \mathbf{b}^{(\ell)}
    \right),
    \label{eq:gtcg_conv}
\end{equation}
where $\mathbf{W}_1^{(\ell)}, \mathbf{W}_2^{(\ell)} \in \mathbb{R}^{d \times d}$ are
learnable projection matrices, $\mathbf{b}^{(\ell)}$ is a bias, and $\mathcal{N}(i)$
denotes the neighbors of agent $i$ in $\tilde{\mathbf{A}}$. The first term transforms
agent $i$'s own embedding; the second aggregates incoming messages from neighbors, each
weighted by an attention coefficient $\alpha_{ij}^{(\ell)}$ computed via scaled
dot-product attention between a query derived from the receiver and a key derived from
the sender:
\begin{equation}
    \alpha_{ij}^{(\ell)} = \frac{
        \exp\!\left(
            \bigl(\mathbf{W}_3^{(\ell)}\, \mathbf{h}_i^{(\ell)}\bigr)^\top
            \bigl(\mathbf{W}_4^{(\ell)}\, \mathbf{h}_j^{(\ell)}\bigr)
            \,/\, \sqrt{d}
        \right)
    }{
        \sum_{k \in \mathcal{N}(i)}
        \exp\!\left(
            \bigl(\mathbf{W}_3^{(\ell)}\, \mathbf{h}_i^{(\ell)}\bigr)^\top
            \bigl(\mathbf{W}_4^{(\ell)}\, \mathbf{h}_k^{(\ell)}\bigr)
            \,/\, \sqrt{d}
        \right)
    },
    \label{eq:gtcg_attention}
\end{equation}
where $\mathbf{W}_3^{(\ell)}$ and $\mathbf{W}_4^{(\ell)}$ are learnable query and key
projections. This formulation connects directly to the three sub-problems of coordination
identified in the introduction. The graph $\tilde{\mathbf{A}}$ determines which
agents can exchange information and modulates the strength of their interactions. The key projection $\mathbf{W}_4$ controls what
each sender makes visible. The query projection $\mathbf{W}_3$ controls how each
receiver filters incoming signals in light of its own current state. Because
$\alpha_{ij}^{(\ell)}$ depends on both $\mathbf{h}_i^{(\ell)}$ and
$\mathbf{h}_j^{(\ell)}$, the same neighbor may contribute different information to
different receivers, and to the same receiver at different timesteps; the messages are
content-dependent, asymmetric, and learned end-to-end. The $L$ layers are applied
sequentially, so after $L$ steps each agent's representation has been informed by agents
up to $L$ hops away, enabling indirect coordination through intermediate agents. We use
$L = 2$ in all experiments.

\paragraph{Preserving local information via residual connection}
A risk of iterative message passing is that locally observed information may be diluted or
overwritten by coordination signals. To prevent this, we add a residual connection from initial embeddings to output of the final graph transformer layer:
\begin{equation}
    \tilde{\mathbf{h}}_i = \mathbf{h}_i^{(0)} + \mathbf{h}_i^{(L)}.
    \label{eq:residual}
\end{equation}
This ensures that $\tilde{\mathbf{h}}_i$ retains agent $i$'s full local information, with
the graph transformer layers contributing an additive coordination refinement. The design
reflects a deliberate principle: coordination should complement local knowledge, not
replace it.

\paragraph{From enriched representations to action values}
The coordination-enriched embeddings $\tilde{\mathbf{h}}_i$ are passed through a shared
aggregation network $\phi_{\text{agg}}$ (a two-layer MLP with the same architecture as
$\phi_{\text{enc}}$), producing final per-agent representations $\mathbf{z}_i =
\phi_{\text{agg}}(\tilde{\mathbf{h}}_i) \in \mathbb{R}^{d}$. These are processed by a
shared agent network $\psi$ (a three-layer feedforward network with hidden dimension
$d_a = 64$) that outputs individual action values:
\begin{equation}
    Q_i(\tau_i, \cdot\,) = \psi(\mathbf{z}_i) \in \mathbb{R}^{|\mathcal{A}_i|}.
    \label{eq:q_values}
\end{equation}
At execution time, each agent independently selects $a_i = \arg\max_{a}\, Q_i(\tau_i, a)$.
So, as in prior coordination graph methods (e.g., \cite{guestrin2002coordinated, bohmer2020deep, li2020deepimplicit, wang2022contextaware, kang2022non, duan2024group}), no explicit inter-agent 
communication protocol is required at execution time; coordination is realized implicitly 
within the shared model during the forward pass. 
% No inter-agent communication occurs at this stage: the coordination has already been
% resolved inside each agent's representation during the forward pass, enabling fully
% decentralized execution.

\paragraph{Centralized training with value mixing}
For centralized training, the individual Q-values are combined via a QMIX mixing
network~\citep{rashid2020monotonic} to produce the joint value estimate:
\begin{equation}
    Q_{\text{tot}}(\boldsymbol{\tau}, \mathbf{a}, s) =
    f_{\text{mix}}\!\left(Q_1(\tau_1, a_1), \dots, Q_n(\tau_n, a_n);\, s\right),
    \label{eq:mixing}
\end{equation}
where $f_{\text{mix}}$ is a feedforward network whose non-negative weights are generated
by hypernetworks conditioned on the global state $s$. The monotonicity constraint
$\partial Q_{\text{tot}} / \partial Q_i \geq 0$ ensures that the joint greedy action
decomposes into per-agent greedy actions, maintaining consistency between centralized
training and decentralized execution. The entire architecture, including the encoder, attention module,
graph transformer layers, aggregator, agent network, and mixer, is trained end-to-end by
minimizing the TD loss with double Q-learning~\citep{van2016deep}:
\begin{equation}
    \mathcal{L}(\theta) = \mathbb{E}_{\mathcal{B}}\!\left[
        \Bigl(
            r + \gamma\, Q_{\text{tot}}^{-}\!\bigl(\boldsymbol{\tau}', \mathbf{a}^*, s'\bigr)
            - Q_{\text{tot}}(\boldsymbol{\tau}, \mathbf{a}, s)
        \Bigr)^2
    \right],
    \label{eq:loss}
\end{equation}
where $\mathcal{B}$ is a replay buffer of collected episodes, $Q_{\text{tot}}^{-}$ is a
periodically-updated target network, and $\mathbf{a}^* =
\arg\max_{\mathbf{a}'} Q_{\text{tot}}(\boldsymbol{\tau}', \mathbf{a}', s')$ is selected
by the online network. Rewards are standardized using a running mean and variance estimate
to stabilize training. Exploration follows $\epsilon$-greedy action selection with linear $\epsilon$-annealing. 

% \setlength{\intextsep}{8pt}
% \vspace{-8pt} 
\begin{algorithm}[t]
\caption{The \algo\ Algorithm}
\label{alg:sachi}
\begin{algorithmic}[1]
\Require Observations $\{o_i^t\}_{i=1}^n$, global state $s^t$, agent identifiers
         $\{\mathbf{e}_i\}$, replay buffer $\mathcal{B}$, coordination graph
         $\mathbf{A}$
\Ensure Per-agent actions $\{a_i\}$; updated parameters $\theta$

\Statex \textbf{--- Forward Pass All Agents ---}
\For{each agent $i \in \mathcal{N}$}
    \State $\mathbf{x}_i \leftarrow [o_i^t \;\|\; \mathbf{e}_i]$
           \Comment{Augment observation with agent ID}
    \State $\mathbf{h}_i^{(0)} \leftarrow \phi_{\text{enc}}(\mathbf{x}_i)$
           \Comment{Shared MLP encoder}
\EndFor
\State $\mathbf{G} \leftarrow \mathrm{softmax}\!\left(
        \phi_{\text{att}}(\mathbf{H}^{(0)})\,(\mathbf{H}^{(0)})^\top\right)$
       \Comment{Soft-attention graph modulation}
\State $\tilde{\mathbf{A}} \leftarrow \mathbf{A} \odot \mathbf{G}$
       \Comment{Modulated adjacency}
\For{$\ell = 0, \dots, L-1$}
    \For{each agent $i \in \mathcal{N}$}
        \State Compute $\alpha_{ij}^{(\ell)}$ via Eq.~\eqref{eq:gtcg_attention}
               for all $j \in \mathcal{N}(i)$
        \State $\mathbf{h}_i^{(\ell+1)} \leftarrow$ Eq.~\eqref{eq:gtcg_conv}
               \Comment{Graph transformer update}
    \EndFor
\EndFor
\For{each agent $i \in \mathcal{N}$}
    \State $\tilde{\mathbf{h}}_i \leftarrow \mathbf{h}_i^{(0)} + \mathbf{h}_i^{(L)}$
           \Comment{Residual connection}
    \State $\mathbf{z}_i \leftarrow \phi_{\text{agg}}(\tilde{\mathbf{h}}_i)$
           \Comment{Shared aggregator}
    \State $Q_i(\tau_i, \cdot) \leftarrow \psi(\mathbf{z}_i)$
           \Comment{Per-agent action values}
    \State $a_i \leftarrow \arg\max_{a}\, Q_i(\tau_i, a)$
           \Comment{Decentralized action selection}
\EndFor

\Statex \textbf{--- Training Step ---}
\State Sample minibatch $(\boldsymbol{\tau}, \mathbf{a}, r, \boldsymbol{\tau}', s, s')
       \sim \mathcal{B}$
\State $Q_{\text{tot}} \leftarrow f_{\text{mix}}(Q_1, \dots, Q_n;\, s)$
       \Comment{QMIX value mixing}
\State $\mathbf{a}^* \leftarrow \arg\max_{\mathbf{a}'} Q_{\text{tot}}(\boldsymbol{\tau}',
       \mathbf{a}', s')$
\State Compute $\mathcal{L}(\theta)$ via Eq.~\eqref{eq:loss} and update $\theta$
       \Comment{TD loss, double Q-learning}
\State Periodically update target network $Q_{\text{tot}}^{-} \leftarrow Q_{\text{tot}}$
\end{algorithmic}
\end{algorithm}

\paragraph{Design rationale}
The architecture of \algo\ makes three deliberate design choices that together address the
information-integration bottleneck. First, the soft-attention modulation layer provides a
global, input-dependent reweighting of the coordination graph before message passing
begins, allowing the model to suppress structurally irrelevant agent pairs at low
computational cost. Second, and more critically, the graph transformer layers provide
content-dependent, receiver-sensitive message passing: because $\alpha_{ij}^{(\ell)}$
depends on both the sender's and receiver's current embeddings, each agent selectively
accumulates precisely the coordination-relevant signals it needs from its neighbors, rather
than aggregating all available information indiscriminately. This is the sense in which the
information integration is \emph{holistic}, i.e., not a collection of everything, but an
intelligent, state-conditioned distillation of what matters. Third, the residual connection
ensures that this coordination signal is additive rather than substitutive, preserving the
integrity of each agent's local observation throughout the enrichment process. 

\section{Experimental Design}
\label{sec:experiments}
We evaluate \algo\ on a suite of five cooperative multi-agent tasks to
collectively probe the full spectrum of coordination demands that arise in
cooperative MARL: spatial symmetry-breaking, communicative grounding,
role-asymmetric signaling, private protocol learning under adversarial pressure,
and adversarially contested spatial coverage. Together, these environments
stress-test the three core sub-problems of action coordination that \algo\ is
designed to address: which agents must exchange information, what information
they should transmit, and how each receiver should integrate incoming signals
given its own state, while varying the structure of the coordination bottleneck
across tasks. We describe each environment, justify its inclusion, detail the
baselines against which \algo\ is evaluated, and specify the experimental
protocol used to ensure fair and reproducible comparisons.

\subsection{Environments}
\label{sec:environments}

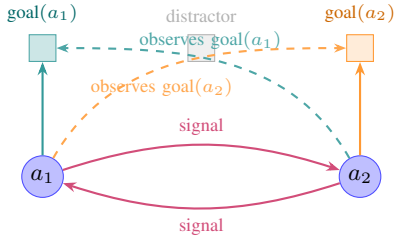
\begin{figure}[h]
\centering
\begin{tikzpicture}[
    agent/.style={circle, draw=blue!70, fill=blue!25,
                  minimum size=0.55cm, inner sep=0pt, font=\small\bfseries},
    landmark/.style={regular polygon, regular polygon sides=4,
                     draw=gray!60, fill=gray!12,
                     minimum size=0.50cm, inner sep=0pt},
    tgtA/.style={regular polygon, regular polygon sides=4,
                 draw=teal!70, fill=teal!20,
                 minimum size=0.50cm, inner sep=0pt},
    tgtB/.style={regular polygon, regular polygon sides=4,
                 draw=orange!80, fill=orange!15,
                 minimum size=0.50cm, inner sep=0pt},
    obs/.style={-{Stealth[length=4.5pt]}, dashed, thick},
    comm/.style={-{Stealth[length=5pt]}, purple!70, thick},
    nav/.style={-{Stealth[length=5pt]}, thick},
    lbl/.style={font=\scriptsize}
]
\node[agent] (A1) at (0.0, 1.3) {$a_1$};
\node[agent] (A2) at (4.2, 1.3) {$a_2$};
\node[tgtA] (T1) at (0.0, 3.0) {};
\node[lbl, text=teal!80!black, above=1pt of T1] {$\text{goal}(a_1)$};
\node[tgtB] (T2) at (4.2, 3.0) {};
\node[lbl, text=orange!80!black, above=1pt of T2] {$\text{goal}(a_2)$};
\node[landmark] (Ld) at (2.1, 3.0) {};
\node[lbl, text=gray!60, above=1pt of Ld] {distractor};
\draw[obs, teal!70] (A2) to[out=120, in=0]
    node[above, lbl, text=teal!70, pos=0.55] {observes $\text{goal}(a_1)$} (T1);
\draw[obs, orange!70] (A1) to[out=60, in=180]
    node[below, lbl, text=orange!70, pos=0.45] {observes $\text{goal}(a_2)$} (T2);
\draw[comm] (A1) to[bend left=18]
    node[above, lbl, text=purple!70] {signal} (A2);
\draw[comm] (A2) to[bend left=18]
    node[below, lbl, text=purple!70] {signal} (A1);
\draw[nav, teal!80]   (A1) -- (T1);
\draw[nav, orange!80] (A2) -- (T2);
\end{tikzpicture}
\caption{\textsc{Reference}: $a_1$ observes $a_2$'s goal (orange dashed)
  and vice versa (teal dashed). Each agent signals this (purple) so
  its partner can navigate to the correct landmark (solid arrows).}
\label{fig:diag_reference}
\end{figure} 

\paragraph{\textsc{Reference}~\cite{lowe2017multi}} is a cooperative navigation task in which two agents must each reach a distinct target landmark, but neither observes its own target. Instead, each agent observes its partner’s goal and must communicate this through its actions. Success requires learning a shared protocol in which each agent acts as both an informative sender and a faithful receiver. The task directly tests content-dependent coordination, as agents must extract and transmit task-relevant information between complementary observations (Figure~\ref{fig:diag_reference}).

\begin{figure}[h]
\centering
\begin{tikzpicture}[
    anode/.style={rectangle, rounded corners=4pt,
                  minimum width=1.0cm, minimum height=0.6cm,
                  font=\scriptsize\bfseries},
    alice/.style={anode, draw=teal!70,   fill=teal!15},
    bob/.style  ={anode, draw=blue!70,   fill=blue!15},
    eve/.style  ={anode, draw=red!70,    fill=red!15},
    chan/.style ={rectangle, draw=gray!50, fill=gray!8,
                  minimum width=1.1cm, minimum height=0.55cm,
                  font=\scriptsize, rounded corners=2pt},
    msg/.style ={-{Stealth[length=5pt]}, thick},
    lbl/.style ={font=\scriptsize}
]
\node[alice] (Al) at (0,   0) {Alice};
\node[chan]  (Ch) at (2.3, 0) {channel};
\node[bob]   (Bo) at (4.6, 0) {Bob};
\node[eve]   (Ev) at (2.3,-1.4) {Eve};
\draw[msg, teal!80]  (Al) -- node[above, lbl]{$\phi(m)$} (Ch);
\draw[msg, blue!80]  (Ch) -- node[above, lbl]{$\hat{m}$} (Bo);
\draw[msg, red!70, dashed] (Ch) -- node[right, lbl, text=red!70]{intercept} (Ev);
\node[lbl, text=teal!80!black] at (0,   -0.65) {$r^+$ if Bob correct};
\node[lbl, text=red!70]        at (2.3, -2.1)  {$r^+$ if Eve correct};
\node[lbl, text=blue!70]       at (4.6, -0.65) {$\hat{m} = m$\,?};
\node[lbl] at (2.3, 0.62) {\textbf{\large\color{teal!70}$\oslash$}};
\node[lbl, text=gray!60, font=\tiny] at (2.3, 0.95) {private code};
\end{tikzpicture}
\caption{\textsc{Crypto}: Alice encodes $m$ as $\phi(m)$ over a shared
  channel. Bob must reconstruct $m$; Eve must not.}
\label{fig:diag_crypto}
\end{figure}
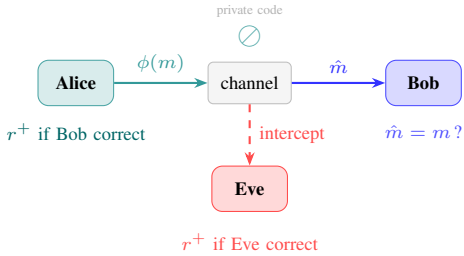 

\paragraph{\textsc{Crypto}~\cite{lowe2017multi}} casts coordination as a communication problem under adversarial pressure. Alice encodes a message for Bob over a shared channel, while an adversary (Eve) attempts to decode it. The cooperative pair is rewarded when Bob succeeds and Eve fails. This requires learning signals that are informative to teammates but uninformative to adversaries, testing selective information encoding and filtering (Figure~\ref{fig:diag_crypto}). 

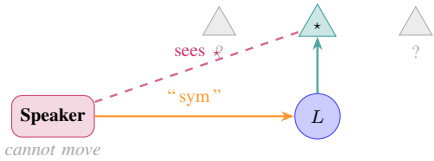
\begin{figure}[h]
\centering
\begin{tikzpicture}[
    speaker/.style={rectangle, rounded corners=4pt,
                    draw=purple!70, fill=purple!15,
                    minimum width=0.9cm, minimum height=0.55cm,
                    font=\scriptsize\bfseries},
    listener/.style={circle, draw=blue!70, fill=blue!20,
                     minimum size=0.6cm, font=\scriptsize\bfseries},
    landmark/.style={regular polygon, regular polygon sides=3,
                     draw=gray!60, fill=gray!15,
                     minimum size=0.5cm, inner sep=1pt, font=\tiny},
    target/.style={regular polygon, regular polygon sides=3,
                   draw=teal!70, fill=teal!20,
                   minimum size=0.5cm, inner sep=1pt, font=\tiny},
    bcast/.style={-{Stealth[length=5pt]}, orange!80, thick},
    nav/.style={-{Stealth[length=5pt]}, teal!70, thick},
    lbl/.style={font=\scriptsize}
]
\node[speaker] (SP) at (0, 1.1) {Speaker};
\node[lbl, text=gray!70] at (0, 0.65) {\textit{cannot move}};
\node[listener] (LI) at (3.5, 1.1) {$L$};
\node[landmark] (La) at (2.2, 2.3) {};
\node[target]   (Lb) at (3.5, 2.3) {\tiny$\star$};
\node[landmark] (Lc) at (4.8, 2.3) {};
\draw[dashed, purple!60, thick] (SP) -- node[above, lbl, text=purple!80]{sees \tiny$\star$} (Lb);
\draw[bcast] (SP) -- node[above, lbl, text=orange!80]{``\,sym\,''} (LI);
\draw[nav] (LI) -- (Lb);
\node[lbl, text=gray!60] at (2.2, 1.95) {?};
\node[lbl, text=gray!60] at (4.8, 1.95) {?};
\end{tikzpicture}
\caption{\textsc{Speaker Listener}: speaker observes target
  ($\star$) and broadcasts a discrete symbol (orange); listener interprets
  the symbol and navigates (teal), unable to see the target directly.}
\label{fig:diag_speaker}
\end{figure} 

\paragraph{\textsc{Speaker Listener}~\cite{lowe2017multi}} introduces fixed role asymmetry. The speaker observes the target but cannot move, while the listener can move but cannot observe the target. The speaker must broadcast a discrete signal encoding the target, which the listener interprets to navigate correctly. Unlike \textsc{Reference}, the communication structure is asymmetric and fixed, testing whether coordination mechanisms can specialize to structured information flow (Figure~\ref{fig:diag_speaker}). 

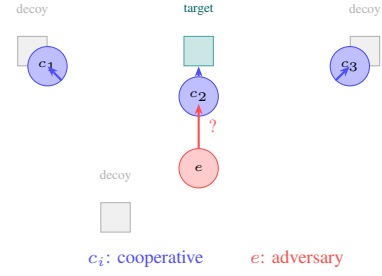
\begin{figure}[h]
\centering
\begin{tikzpicture}[
    coop/.style={circle, draw=blue!70, fill=blue!25,
                 minimum size=0.52cm, inner sep=0pt, font=\tiny\bfseries},
    adv/.style ={circle, draw=red!70,  fill=red!20,
                 minimum size=0.52cm, inner sep=0pt, font=\tiny\bfseries},
    lm/.style  ={regular polygon, regular polygon sides=4,
                 draw=gray!60, fill=gray!12,
                 minimum size=0.55cm, inner sep=0pt},
    target/.style={regular polygon, regular polygon sides=4,
                   draw=teal!70, fill=teal!20,
                   minimum size=0.55cm, inner sep=0pt},
    move/.style={-{Stealth[length=4pt]}, thick},
    lbl/.style ={font=\scriptsize}
]
\node[target] (T)  at (2.2, 2.2) {};
\node[lm]     (L1) at (0.0, 2.2) {};
\node[lm]     (L2) at (4.4, 2.2) {};
\node[lm]     (L3) at (1.1, 0.0) {};
\node[lbl, text=teal!80!black] at (2.2, 2.75) {\tiny target};
\node[coop] (C1) at (0.2, 2.0) {$c_1$};
\node[coop] (C2) at (2.2, 1.6) {$c_2$};
\node[coop] (C3) at (4.2, 2.0) {$c_3$};
\draw[move, blue!70] (C1) -- (L1);
\draw[move, blue!70] (C2) -- (T);
\draw[move, blue!70] (C3) -- (L2);
\node[adv] (Adv) at (2.2, 0.65) {$e$};
\draw[move, red!70] (Adv) -- node[right, lbl, text=red!60]{?} ($(T)!0.45!(Adv)$);
\node[lbl, text=gray!60] at (0.0, 2.75) {\tiny decoy};
\node[lbl, text=gray!60] at (4.4, 2.75) {\tiny decoy};
\node[lbl, text=gray!60] at (1.1, 0.55) {\tiny decoy};
\node[lbl, text=blue!70]  at (1.5, -0.55) {$c_i$: cooperative};
\node[lbl, text=red!70]   at (3.5, -0.55) {$e$: adversary};
\end{tikzpicture}
\caption{\textsc{Adversary}: cooperative agents ($c_i$, blue) distribute
  across all landmarks to conceal true target (teal) from adversary
  ($e$, red), who must infer target from team's configuration.}
\label{fig:diag_adversary}
\end{figure}  

\paragraph{\textsc{Adversary}~\cite{lowe2017multi}} is a mixed cooperative-competitive task in which a team of agents must reach a hidden target while preventing a faster adversary from identifying it. The cooperative agents are rewarded for occupying the target, while the adversary is rewarded for reaching it. Effective strategies require distributing across landmarks to conceal the target, forcing agents to coordinate both coverage and deception under an active opponent (Figure~\ref{fig:diag_adversary}). 

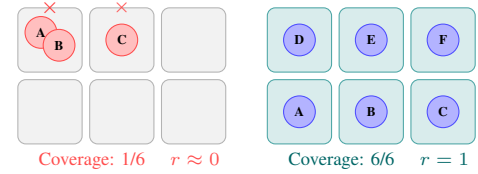
\begin{figure}[h]
\centering
\begin{tikzpicture}[
    zone/.style={draw=gray!60, fill=gray!10, rounded corners=3pt,
                 minimum width=0.85cm, minimum height=0.85cm},
    zonefull/.style={draw=teal!70, fill=teal!15, rounded corners=3pt,
                     minimum width=0.85cm, minimum height=0.85cm},
    agent/.style={circle, draw=blue!70, fill=blue!30,
                  minimum size=0.42cm, inner sep=0pt, font=\tiny\bfseries},
    badagent/.style={circle, draw=red!70, fill=red!25,
                     minimum size=0.42cm, inner sep=0pt, font=\tiny\bfseries},
    label/.style={font=\scriptsize, align=center}
]
% \node[label] at (1.05, 2.55) {\textbf{Without coordination}};
\foreach \col in {0,1,2} {
  \foreach \row in {0,1} { \node[zone] at (\col*0.95, \row*0.95) {}; }
}
\node[badagent] at (-0.12, 1.05) {A};
\node[badagent] at ( 0.12, 0.88) {B};
\node[badagent] at ( 0.95, 0.95) {C};
\node[font=\footnotesize, text=red!80] at (0.0,  1.38) {$\times$};
\node[font=\scriptsize,   text=red!70] at (0.95, 1.38) {$\times$};
\node[label, text=red!70] at (1.05, -0.65) {Coverage: 1/6 \quad $r \approx 0$};
\begin{scope}[xshift=3.3cm]
% \node[label] at (1.05, 2.55) {\textbf{With \algo}};
\foreach \col in {0,1,2} {
  \foreach \row in {0,1} { \node[zonefull] at (\col*0.95, \row*0.95) {}; }
}
\node[agent] at (0.00, 0.00) {A};
\node[agent] at (0.95, 0.00) {B};
\node[agent] at (1.90, 0.00) {C};
\node[agent] at (0.00, 0.95) {D};
\node[agent] at (0.95, 0.95) {E};
\node[agent] at (1.90, 0.95) {F};
\node[label, text=teal!80!black] at (1.05, -0.65) {Coverage: 6/6 \quad $r = 1$};
\end{scope}
\end{tikzpicture}
\caption{\textsc{Disperse}: Agents must try to achieve full coverage.}
\label{fig:diag_disperse}
\end{figure} 

\paragraph{\textsc{Disperse}~\cite{wang2022contextaware}} requires $n$ agents to occupy $n$ identical resource zones with exactly one agent per zone. Rewards depend on conflict-free coverage, but all zones are locally indistinguishable, making symmetry-breaking the central challenge. Greedy local policies lead to clustering and under-coverage. Solving the task requires agents to incorporate teammates’ implicit intentions into their representations to achieve globally consistent assignments (Figure~\ref{fig:diag_disperse}). 

% \textbf{Why this suite?}
These environments span complementary coordination regimes. Reference and Speaker Listener evaluate learned communication under symmetric and asymmetric roles. Crypto adds adversarial pressure to communication, requiring selective information sharing. Adversary tests spatial coordination under competition, while Disperse isolates symmetry-breaking in purely cooperative settings. Together, they provide a balanced evaluation of \algo's ability to integrate information across agents under diverse coordination structures.

\subsection{Baselines}
\label{sec:baselines} 

We compare \algo\ against twelve baselines spanning a spectrum from no coordination to explicit graph-based representation learning. This ensures that performance differences reflect \algo's architectural contribution rather than generic coordination effects. 

\paragraph{No coordination}
\textbf{IQL}~\cite{tan1993multi} and \textbf{IPPO}~\cite{yu2022surprising} treat each agent independently, conditioning only on local observations. These methods provide lower bounds, quantifying performance in the absence of coordination. 

\paragraph{Implicit coordination}
These methods induce coordination through the training objective without explicit inter-agent communication. \textbf{VDN}~\cite{sunehag2017value} decomposes the joint value as a sum of per-agent utilities, while \textbf{QMIX}~\cite{rashid2020monotonic} generalizes this to a monotonic nonlinear mixing conditioned on global state. \textbf{QTRAN}~\cite{son2019qtran}, \textbf{QPLEX}~\cite{wang2020qplex} further relax these structural constraints to represent broader classes of joint value functions, at the cost of additional complexity. \textbf{MAPPO}~\cite{yu2022surprising} uses a centralized critic during training but executes decentralized policies. This group represents the dominant paradigm in cooperative MARL and forms the primary category for evaluating \algo. 

% \paragraph{Implicit coordination}
% These methods induce coordination through the training objective without explicit inter-agent communication. \textbf{VDN}~\cite{sunehag2017value} decomposes the joint value as a sum of per-agent utilities, while \textbf{QMIX}~\cite{rashid2020monotonic} generalizes this to a monotonic nonlinear mixing conditioned on global state. \textbf{MAPPO}~\cite{yu2022surprising} uses a centralized critic during training but executes decentralized policies. This group represents the dominant paradigm in cooperative MARL and forms the primary baseline for evaluating \algo.

\paragraph{Explicit joint behavior modeling}
\textbf{FOP}~\cite{zhang2021fop} models the joint policy distribution directly under a maximum-entropy objective, capturing coordination at the level of joint actions. It provides a comparison point for explicit joint reasoning beyond additive or monotonic structures.

\paragraph{Coordination graphs}
\textbf{DCG} \cite{bohmer2020deep} models pairwise dependencies via a fully connected coordination graph with learned edge payoffs, optimized through message passing. \textbf{CASEC} \cite{wang2022contextaware} extends this by learning sparse, context-dependent coordination structures. Both encode coordination in the value function via pairwise interactions, contrasting with \algo's representation-level approach.

\paragraph{Graph representation learning}
These methods are closest to \algo. \textbf{DGN}~\cite{Jiang2020Graph} uses graph convolution with uniform aggregation, treating all neighbors equally. \textbf{DICG}~\cite{li2020deepimplicit} learns a dynamic interaction graph but employs sender-only attention, where message importance depends only on the sender. Comparing against this group isolates the effect of \algo's core components. 

% \textbf{Why this baseline set?}
Each group provides a distinct counterfactual. Groups I--II establish the gap between independent learning and implicit coordination. Group III tests explicit joint policy modeling. Group IV evaluates pairwise coordination in the value function. Group V provides the closest architectural comparison, isolating whether receiver-dependent attention improves over uniform aggregation (DGN) and sender-only attention with dynamic topology (DICG). Together, these baselines span key dimensions of cooperative MARL: learning paradigm, coordination mechanism, and graph structure, ensuring a controlled and comprehensive evaluation of \algo.

\subsection{Evaluation Protocol}
\label{sec:protocol}

All algorithms are trained for up to $2 \times 10^6$ environment steps across
all tasks. All implementations share a common
training pipeline, including identical replay buffer capacity, minibatch size, target
network update frequency, optimizer (Adam), and learning rate schedule, to
ensure that observed performance differences are attributable to algorithmic
design rather than implementation variance. For Q-learning methods, exploration
follows $\epsilon$-greedy action selection with $\epsilon$ annealed linearly
from $1.0$ to $0.05$ over the first $50{,}000$ timesteps; for policy-gradient
methods, entropy regularization is used with schedules matched across
comparable algorithms. Rewards are standardized online using a running mean and
variance estimate for all methods to stabilize training across environments with
different reward scales. 
Performance is reported as mean episodic reward over five independent
trials, with standard deviation shown as a shaded region surrounding
the learning curve throughout training. We evaluate two complementary
metrics: final performance, defined as the mean reward averaged over the
last $10\%$ of training steps, which measures the quality of the converged
policy; and sample efficiency, measured as the normalized area under the
learning curve, which captures how rapidly each method approaches its
asymptotic performance. 
Hyperparameters for all baselines follow their original implementations, including any reported environment-specific tuning. \algo\ uses $L=2$, $d=64$, and single-head attention ($K=1$). All methods are run under a unified experimental setup \cite{papoudakis2021benchmarking,samvelyan19smac,hu2021rethinking} to ensure consistent and comparable evaluation. 
% Statistical
% significance of pairwise comparisons between \algo\ and each baseline on final
% performance is assessed via a two-sided Welch's $t$-test across seeds, with
% significance threshold $\alpha = 0.05$. 

% Hyperparameters for all baselines are set to the values reported in their
% respective original publications; where environment-specific hyperparameter
% tuning was performed in the original work, we adopt the published tuned values
% directly. \algo\ uses $L = 2$ graph transformer layers, embedding dimension
% $d = 64$, and single-head attention ($K = 1$) throughout, consistent with
% the ablation analysis reported in Section~\ref{sec:ablations}. All experiments
% are implemented in PyTorch and run on identical hardware to ensure runtime
% comparability across methods.

\section{Results}
\label{sec:results}

We organize the empirical evaluation around a progression of increasingly stringent questions. We begin with the most basic question: ``does \algo\ learn good policies?", and proceed to questions of statistical robustness, efficiency, and architectural necessity. 
% Each subsection raises the bar; \algo\ clears it each time. 

\subsection{Per-Environment Performance}
\label{sec:main_results}

\begin{figure*}[t]
    \centering
    \includegraphics[width=0.85\linewidth]{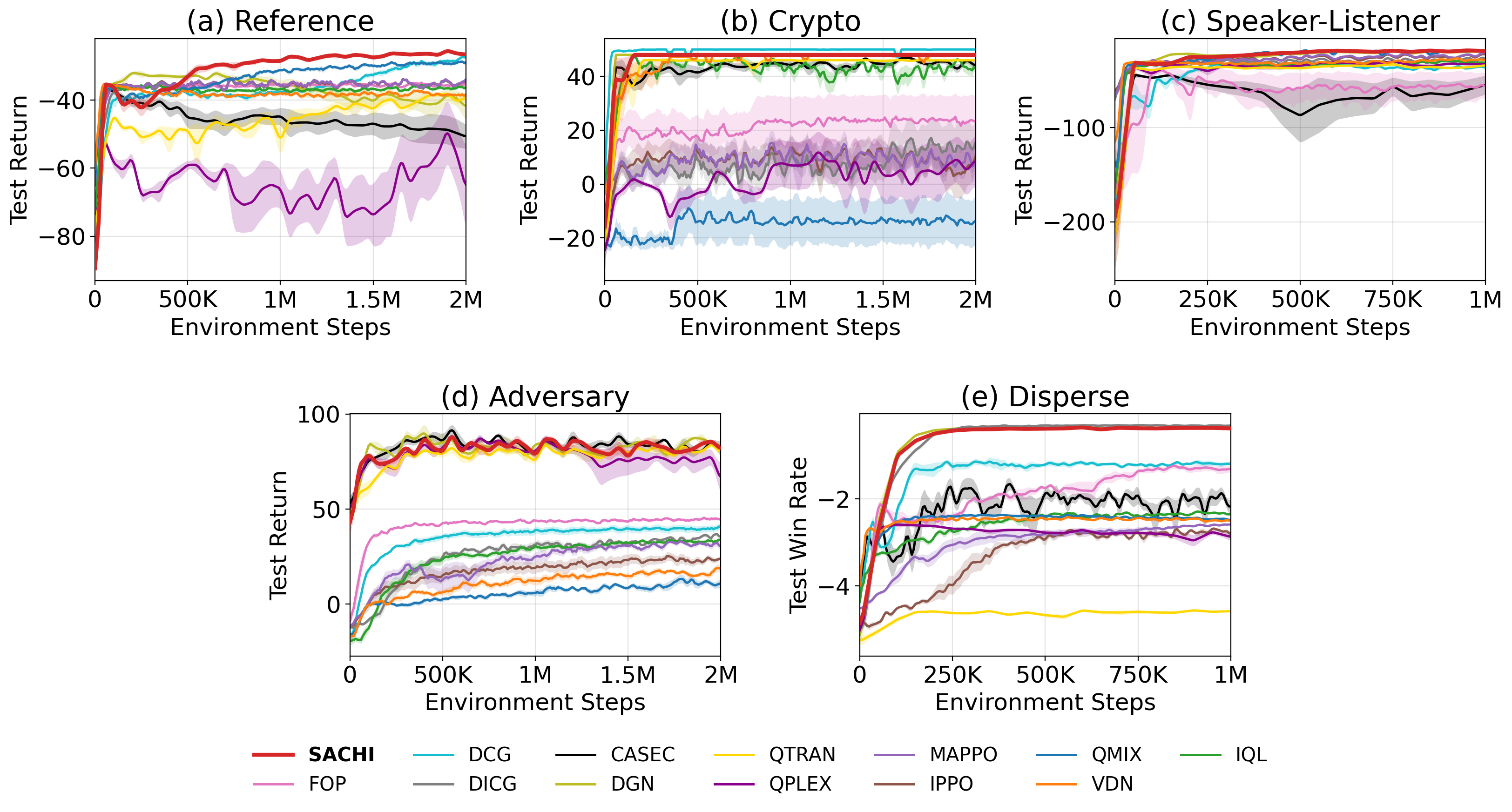}
    \caption{Learning curves across five cooperative environments (mean $\pm$ standard error over five seeds). \algo\ (red, bold) matches or exceeds the best baseline on every environment by convergence.}
    \label{fig:main_results}
% \vspace{-15pt} 
\end{figure*}

% Auto-generated by latex_table_subset.py — do not edit by hand.
\begin{table}[t]
\centering
\small
\setlength{\tabcolsep}{4pt}
\caption{Final performance (mean $\pm$ stderr over seeds, last 2\% of training). Best per environment in bold.}
\label{tab:sachi_subset}
\resizebox{\columnwidth}{!}{
\begin{tabular}{lccccc}
\toprule
Method & Reference & Crypto & Speaker-Listener & Adversary & Disperse \\
\midrule
\textbf{SACHI} & \textbf{-25.39 $\pm$ 0.35} & \underline{48.00 $\pm$ 0.00} & \textbf{-17.22 $\pm$ 0.72} & \textbf{85.04 $\pm$ 2.40} & \underline{-0.37 $\pm$ 0.01} \\
\midrule
QMIX & -28.81 $\pm$ 0.39 & -14.34 $\pm$ 8.83 & -18.45 $\pm$ 0.22 & 10.45 $\pm$ 2.01 & -2.57 $\pm$ 0.02 \\
VDN & -38.44 $\pm$ 0.39 & 47.70 $\pm$ 0.18 & -25.68 $\pm$ 1.73 & 18.69 $\pm$ 1.85 & -2.59 $\pm$ 0.01 \\
IQL & -36.12 $\pm$ 0.33 & 42.38 $\pm$ 1.70 & -27.70 $\pm$ 0.71 & 33.46 $\pm$ 0.78 & -2.36 $\pm$ 0.01 \\
MAPPO & -34.57 $\pm$ 0.07 & 7.31 $\pm$ 5.98 & -19.44 $\pm$ 1.30 & 31.83 $\pm$ 1.61 & -2.54 $\pm$ 0.01 \\
IPPO & -34.69 $\pm$ 0.09 & 7.99 $\pm$ 5.89 & -20.38 $\pm$ 1.24 & 23.54 $\pm$ 1.07 & -2.78 $\pm$ 0.03 \\
FOP & -35.38 $\pm$ 0.35 & 23.69 $\pm$ 9.29 & -48.15 $\pm$ 12.43 & 44.68 $\pm$ 0.30 & -1.12 $\pm$ 0.02 \\
DICG & -34.97 $\pm$ 0.10 & 13.35 $\pm$ 8.70 & -22.45 $\pm$ 1.56 & 36.51 $\pm$ 0.68 & \textbf{-0.36 $\pm$ 0.01} \\
DGN & -41.95 $\pm$ 4.08 & 48.00 $\pm$ 0.00 & -17.82 $\pm$ 0.65 & 83.54 $\pm$ 0.32 & -0.39 $\pm$ 0.01 \\
DCG & -27.34 $\pm$ 0.41 & \textbf{50.00 $\pm$ 0.00} & -21.19 $\pm$ 0.48 & 40.35 $\pm$ 1.59 & -1.16 $\pm$ 0.02 \\
CASEC & -50.71 $\pm$ 3.54 & 43.90 $\pm$ 2.07 & -43.79 $\pm$ 3.67 & 81.55 $\pm$ 4.03 & -2.00 $\pm$ 0.09 \\
QTRAN & -39.33 $\pm$ 1.39 & 46.07 $\pm$ 0.13 & -34.29 $\pm$ 0.62 & 79.79 $\pm$ 2.36 & -4.55 $\pm$ 0.02 \\
QPLEX & -65.72 $\pm$ 12.42 & 8.98 $\pm$ 9.53 & -29.49 $\pm$ 1.78 & 65.69 $\pm$ 8.15 & -2.99 $\pm$ 0.07 \\
\bottomrule
\end{tabular}
}
% \vspace{-16pt} 
% \vspace{-1em} 
\end{table}  

Figure~\ref{fig:main_results} and Table~\ref{tab:sachi_subset} present a full picture, by making it easy to compare final performance and identify per-environment wins (Table~\ref{tab:sachi_subset}), alongside the learning curves, which reveal training dynamics such as stability, convergence speed, and variability across episodes (Figure~\ref{fig:main_results}). We highlight three qualitative patterns that emerge from the data. 

\paragraph{Pattern 1: \algo\ dominates where coordination is hardest} The most striking separations occur on tasks with tight, asymmetric interdependence. On \textsc{Adversary}, \algo\ reaches $85.0 \pm 2.4$, more than doubling the returns of QMIX ($10.5$), VDN ($18.7$), and IPPO ($23.5$). These value-decomposition and policy-gradient methods have no mechanism for one agent to condition its representation on another's state; on a task where the cooperative team must collectively outmaneuver a pretrained adversary, this limitation is fatal. \algo\ and DGN ($83.5$) are the only methods that break $80$, and both use graph-structured inter-agent information exchange. However, \algo\ does so with $5{\times}$ fewer parameters (Section~\ref{sec:efficiency}). QTRAN ($79.8$) and QPLEX ($65.7$) both fall short despite employing richer value factorizations than QMIX, suggesting that more expressive decompositions of the joint value function alone are insufficient without explicit inter-agent information exchange. On \textsc{Reference}, where each agent observes only the other's target and must infer its own, \algo\ converges to $-25.4 \pm 0.4$, ahead of DCG ($-27.3$) and well above the baseline cluster near $-35$. QPLEX collapses to $-65.7 \pm 12.4$ on this task (the worst among all methods) with high variance across seeds. This task is a direct test of the information-integration thesis: an agent literally cannot act well without extracting a specific piece of information from its partner's representation. The graph transformer's query-key mechanism is precisely suited to this extraction. 

\paragraph{Pattern 2: \algo\ matches specialists on saturated tasks} Not every environment has room to differentiate. On \textsc{Crypto}, DCG reaches the ceiling score of $50.0$, with \algo\ and DGN close behind at $48.0$ and QTRAN at $46.1$. On \textsc{Disperse}, DICG ($-0.36$) and \algo\ ($-0.37$) are separated by a margin smaller than either method's standard error, while QTRAN ($-4.55$) and QPLEX ($-2.99$) fall well behind, indicating that the coordination structure of this task is not well captured by value-factorization alone. When a task is solvable by a simpler coordination mechanism, \algo\ does not underperform. It converges to the same region. This is the expected behavior of a strictly more expressive message-passing primitive: it should never do worse than its less expressive counterparts (given sufficient data), and it should do better when the task demands it. 

\paragraph{Pattern 3: Methods that match or beat \algo\ on individual tasks are not uniformly strongest overall}
DCG tops \textsc{Crypto} but achieves only $40.4$ on \textsc{Adversary}, less than half of \algo's score. DICG edges ahead on \textsc{Disperse} but drops to $13.4$ on \textsc{Crypto}. DGN is the strongest overall baseline and remains competitive with \algo\ on several tasks, but its advantage is narrower and less consistent in the aggregate analyses (see Section~\ref{sec:aggregate}). QTRAN performs reasonably on \textsc{Adversary} ($79.8$) and \textsc{Crypto} ($46.1$) but collapses on \textsc{Disperse} ($-4.55$), the worst score of any method on that task. QPLEX shows extreme inconsistency: competitive on \textsc{Adversary} ($65.7$) but the worst performer on \textsc{Reference} ($-65.7$) with standard errors exceeding $12$ points. The practical question is therefore not which method wins a single environment, but which one remains most reliable across tasks. The aggregate results below address this directly. 

\subsection{Aggregate Statistical Analysis}
\label{sec:aggregate}

\begin{figure*}[!htp]
    \centering
    \includegraphics[width=0.85\linewidth]{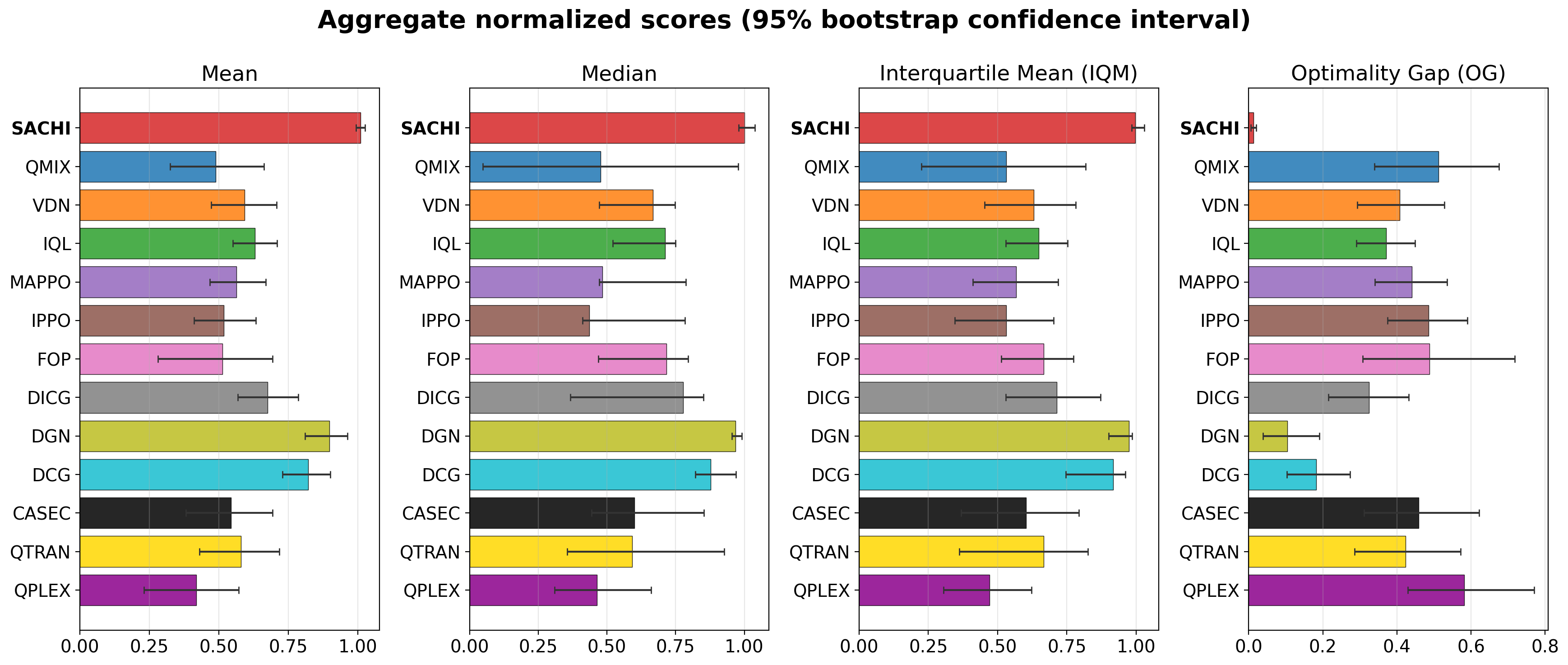}
    \caption{Aggregate normalized scores with $95\%$ bootstrap confidence intervals. \algo\ leads on all four metrics; its CI does not overlap with any baseline on Mean, IQM, or OG.}
    \label{fig:aggregate}
\end{figure*}

Per-environment comparisons are informative but fragile: a method that excels on one task and fails on another can look good or bad depending on which table entry one highlights. Following the recommendations of \cite{agarwal2021deep}, we compute four aggregate metrics over min-max normalized scores with $95\%$ bootstrap confidence intervals. For each method $m$, environment $e$, and seed $s$, let $v_{m,e,s}$ be the mean return over the last $5\%$ of training. We normalize:
\begin{equation}
\hat{v}_{m,e,s} = \frac{v_{m,e,s} - v^{\min}_e}{v^{\max}_e - v^{\min}_e}
\label{eq:normalization}
\end{equation}

\begin{equation}
v^{\min}_e = \min_{m' \in \mathcal{M}_{\text{base}}} \bar{v}_{m',e} \quad, 
\quad v^{\max}_e = \max_{m' \in \mathcal{M}_{\text{base}}} \bar{v}_{m',e}
\label{eq:vmin}
\end{equation}

% \begin{equation}
% v^{\max}_e = \max_{m' \in \mathcal{M}_{\text{base}}} \bar{v}_{m',e}
% \label{eq:vmax}
% \end{equation}
where $\bar{v}_{m',e} = S^{-1}\sum_s v_{m',e,s}$ and $\mathcal{M}_{\text{base}}$ excludes \algo, so that scores can exceed $1.0$.
 
Figure~\ref{fig:aggregate} reports four metrics computed over the pooled set of normalized scores $\{\hat{v}_{m,e,s}\}$ for each method $m$. The Mean and Median have their usual definitions. The Interquartile Mean (IQM) averages only the middle $50\%$ of the score distribution, i.e., it discards the best and worst quartiles, making it robust to both lucky outliers and catastrophic seeds. The Optimality Gap (OG), defined as $\text{OG}(m) = \mathbb{E}[\max(0,\; 1 - \hat{v}_{m,e,s})]$, measures by how much, on average, a method falls short of the best-baseline ceiling ($\hat{v} = 1$); a method that always matches or exceeds the best baseline has $\text{OG} = 0$. For each metric, we construct $95\%$ confidence intervals by bootstrap resampling the (environment $\times$ seed) score pool $2{,}000$ times and taking the $2.5$th and $97.5$th percentiles, requiring no parametric distributional assumptions. Three observations stand out:

\begin{itemize}
    \item \textbf{Consistently at the ceiling.} \algo\ achieves an IQM of $1.00$ with a $95\%$ confidence interval (CI) of $[0.98, 1.03]$. Recall that a normalized score of $1.0$ corresponds to the best baseline's average performance on a given environment. An IQM of $1.00$ means that even after discarding the top and bottom $25\%$ of \algo's runs, the remaining ``typical'' runs match or exceed the best baseline. No other method comes close: next-highest IQM is DGN at ${\approx}0.75$.
 
    \item \textbf{Near-zero shortfall.} \algo's optimality gap is $0.01$ (CI: $[0.01, 0.02]$). To put this in perspective: an OG of $0.01$ means that, on average, \algo\ falls only $1\%$ below the best-baseline ceiling across all environments and seeds. In practical terms, \algo\ almost never underperforms the strongest available alternative on any given task.
 
    \item \textbf{Statistically significant.} \algo's $95\%$ bootstrap CIs do not overlap with those of any baseline on Mean, IQM, or OG. This means the advantage is not an artifact of favorable seed draws or a single dominant environment. 
\end{itemize} 

\begin{figure}[!htp]
    \centering
    \includegraphics[width=0.8\linewidth]{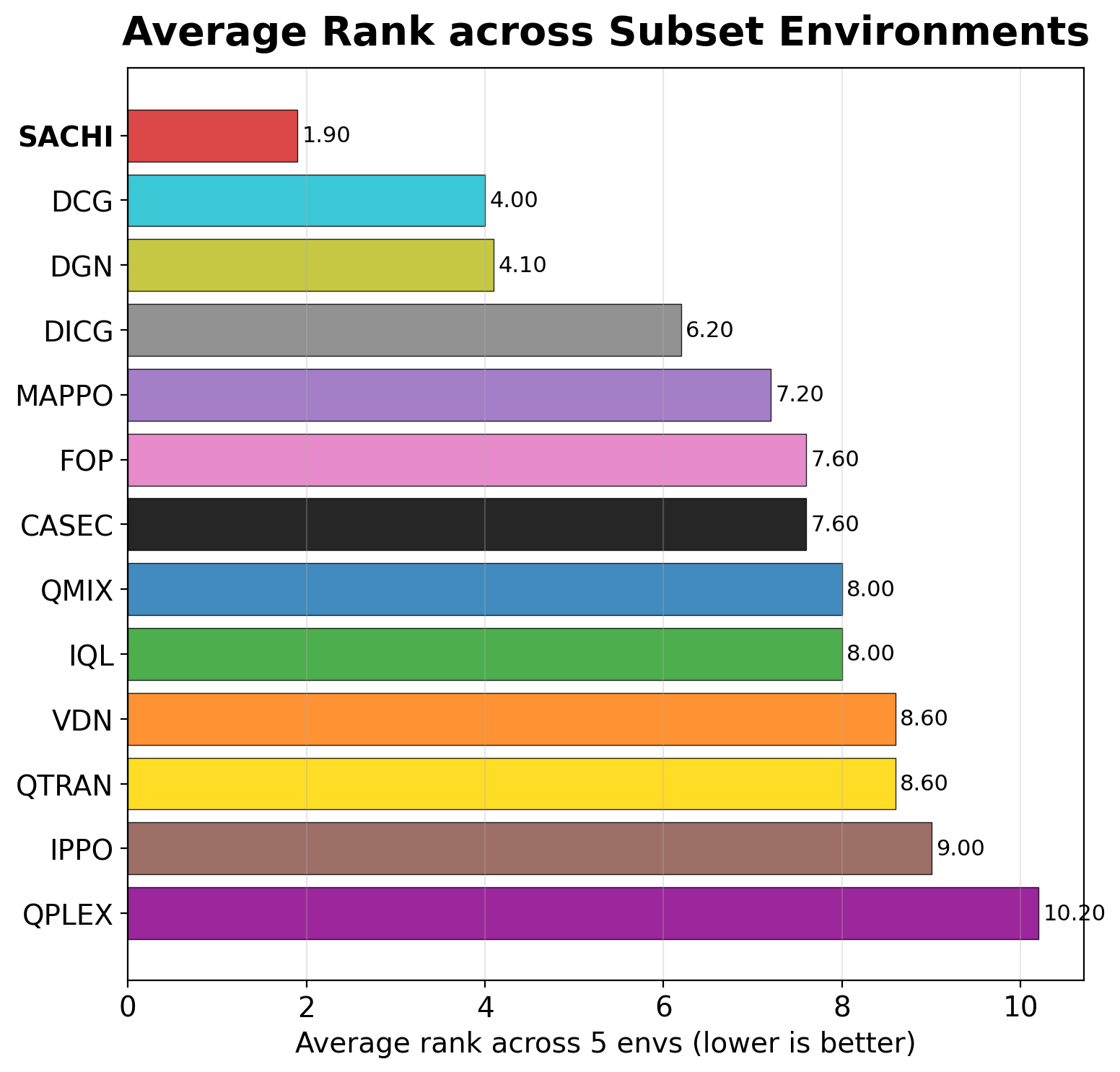}
    \caption{Average rank across five environments (lower is better). \algo\ ranks first or second on every task.}
    \label{fig:rank}
% \vspace{-12pt}
\end{figure}

\textbf{Friedman-style Average Ranks} \cite{demvsar2006statistical}: 
To guard against the possibility that \algo's aggregate score is inflated by one dominant environment, we compute Friedman-style average ranks (Figure~\ref{fig:rank}). For each environment, methods are ranked $1$ (best) to $11$ (worst) by mean final return; ranks are then averaged. \algo\ achieves $1.90$, first or second on every environment. DCG ($3.60$) and DGN ($3.90$) are the nearest competitors, roughly two full ranks behind. The gap to the value-decomposition family (QMIX: $7.20$; VDN: $8.20$; QTRAN: $8.60$; QPLEX: $10.20$) is even larger, spanning more than five ranks. This confirms that \algo's advantage is not an artifact of a single strong showing. 
 
\begin{figure}[!htp]
    \centering
    \includegraphics[width=0.8\linewidth]{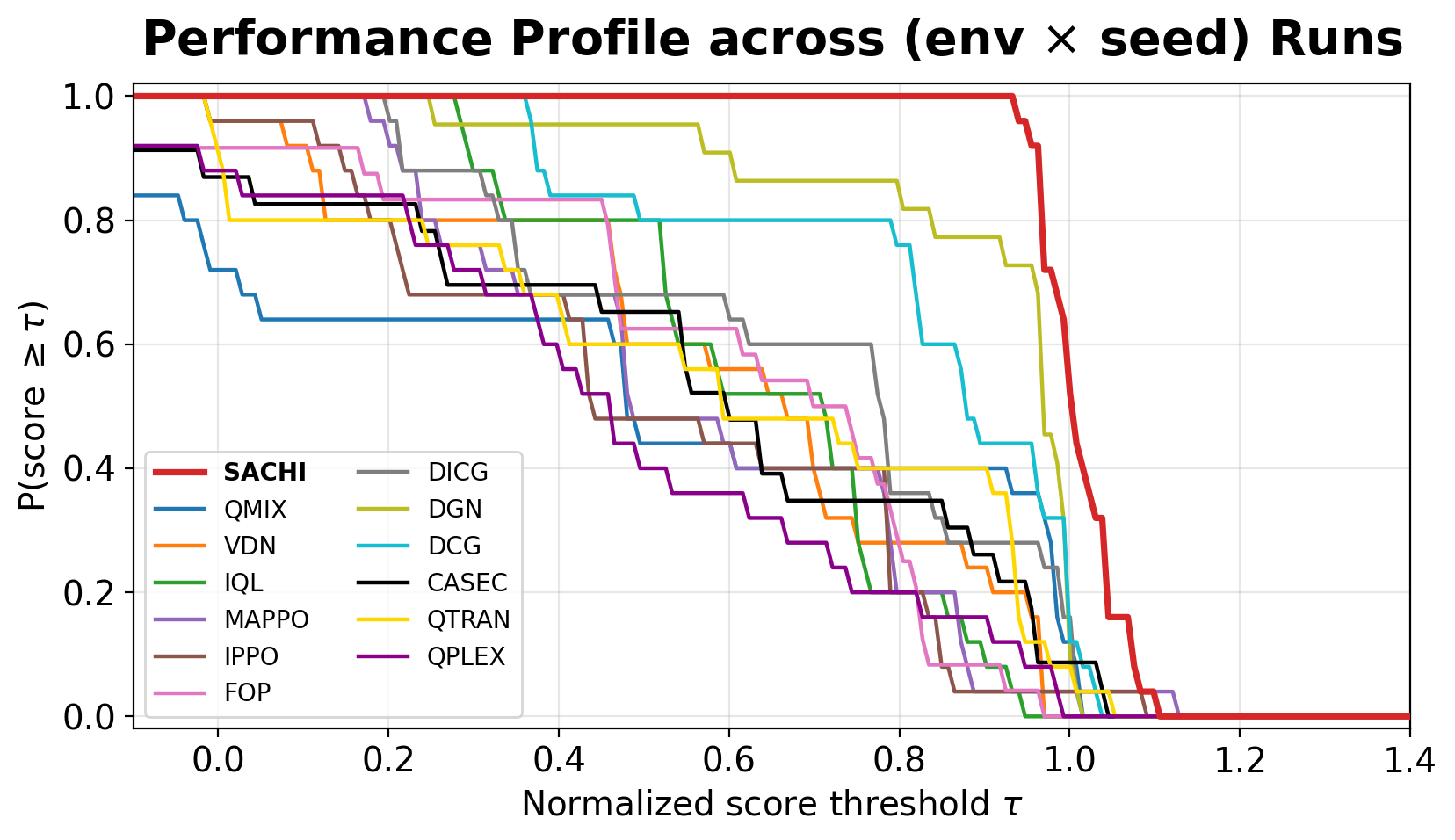}
    \caption{Performance profile. \algo\ stochastically dominates every baseline at every threshold $\tau$.}
    \label{fig:perf_profile}
% \vspace{-20pt}
\end{figure}

\textbf{The Performance Profile}: Figure~\ref{fig:perf_profile} provides the final statistical lens. For each threshold $\tau$, it plots $\Pr(\hat{v}_{m,e,s} \geq \tau)$, i.e., the fraction of (environment $\times$ seed) runs achieving at least score $\tau$. A curve that lies everywhere above another indicates stochastic dominance: the higher method is at least as good on every quantile of the run distribution. \algo's curve dominates all ten baselines at every threshold. At $\tau = 1.0$ (the best-baseline ceiling) over $80\%$ of \algo's runs still exceed this score. DGN, the strongest competitor, drops to ${\approx}40\%$; all other methods fall below $20\%$. Stochastic dominance is a strong property: it implies that \algo\ is preferred under any monotone utility function over normalized scores, not just the specific metrics we report. 

\subsection{Sample Efficiency}
\label{sec:efficiency}
Good final performance means little if it comes at prohibitive cost. We examine efficiency along three axes: sample efficiency over the full trajectory, early-training behavior, and parameter count. 

\begin{figure}[!htp]
    \centering
    \includegraphics[width=0.85\linewidth]{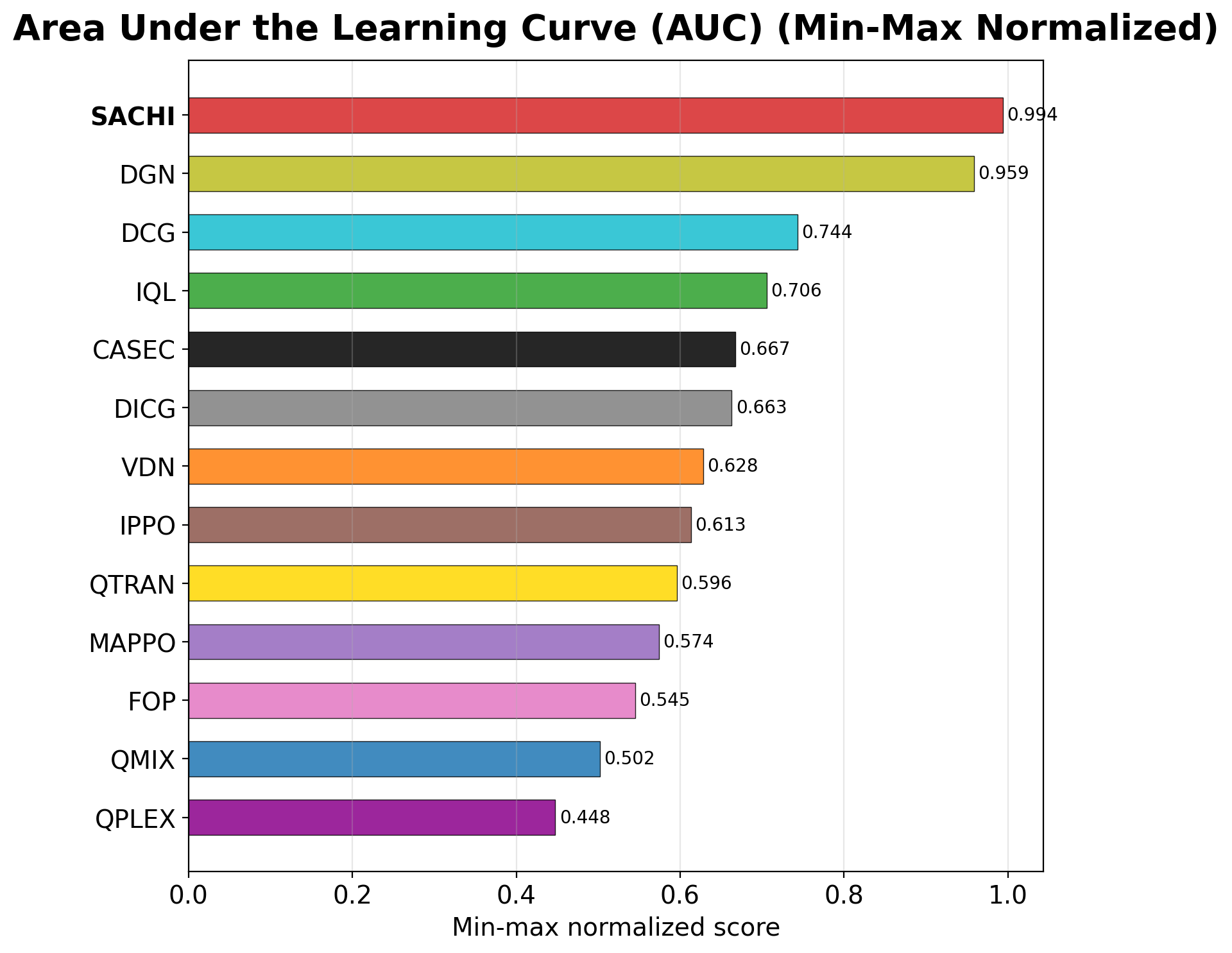}
    \caption{Normalized area under the learning curve (higher is better). \algo\ achieves the highest overall learning efficiency.}
    \label{fig:auc}
% \vspace{-12pt}
\end{figure}

\paragraph{Learning-trajectory efficiency} Let $R_{m,e,s}(t)$ denote the test return of method $m$ on environment $e$ with seed $s$ at training timestep $t$, and let $T$ denote the total training budget. We compute the normalized area under the learning curve:
\begin{equation}
    \text{AUC}_{m,e,s} = \frac{1}{T} \int_0^T R_{m,e,s}(t)\, dt,
    \label{eq:auc}
\end{equation}
estimated via the trapezoidal rule and min-max normalized per environment. Figure~\ref{fig:auc} shows that \algo\ ranks first. Because AUC integrates performance across the entire training trajectory this result confirms that \algo's advantage is not an artifact of late-stage convergence but reflects a consistently stronger learning signal throughout training. 

\paragraph{Jump-start performance} AUC captures the full training trajectory, but it does not reveal how quickly a method begins to coordinate. To isolate early-training behavior, we define the \emph{jump-start value} of method $m$ on environment $e$ as the mean test return over the early-training window:
\begin{equation}
    J(m, e) = \frac{1}{S} \sum_{s=1}^{S} \frac{1}{|\mathcal{T}_s|} \sum_{t \in \mathcal{T}_s} y_s(t),
    \label{eq:jumpstart}
\end{equation}
where $\mathcal{T}_s = \{\, t \in x_s : 20\text{K} \le t \le 100\text{K} \}$, $y_s(t)$ is the test return of seed $s$ at step $t$. The lower bound of $20$K excludes the random-initialization phase, where return magnitude reflects network expressiveness rather than coordination ability. To compare across environments with different reward scales, we normalize each $J(m, e)$ against baseline-only anchors:
\begin{equation}
    \text{score}(m, e) = \frac{J(m, e) - b_{\min}(e)}{b_{\max}(e) - b_{\min}(e)} \times 100, \label{eq:jumpstart_norm_1}
\end{equation}
\begin{equation}
    b_{\min}(e) = \min_{m' \in \mathcal{B}} J(m', e), \quad b_{\max}(e) = \max_{m' \in \mathcal{B}} J(m', e),
    \label{eq:jumpstart_norm_2}
\end{equation}
where $\mathcal{B}$ excludes \algo. 
\begin{figure}[!htp]
    \centering
    \includegraphics[width=0.8\linewidth]{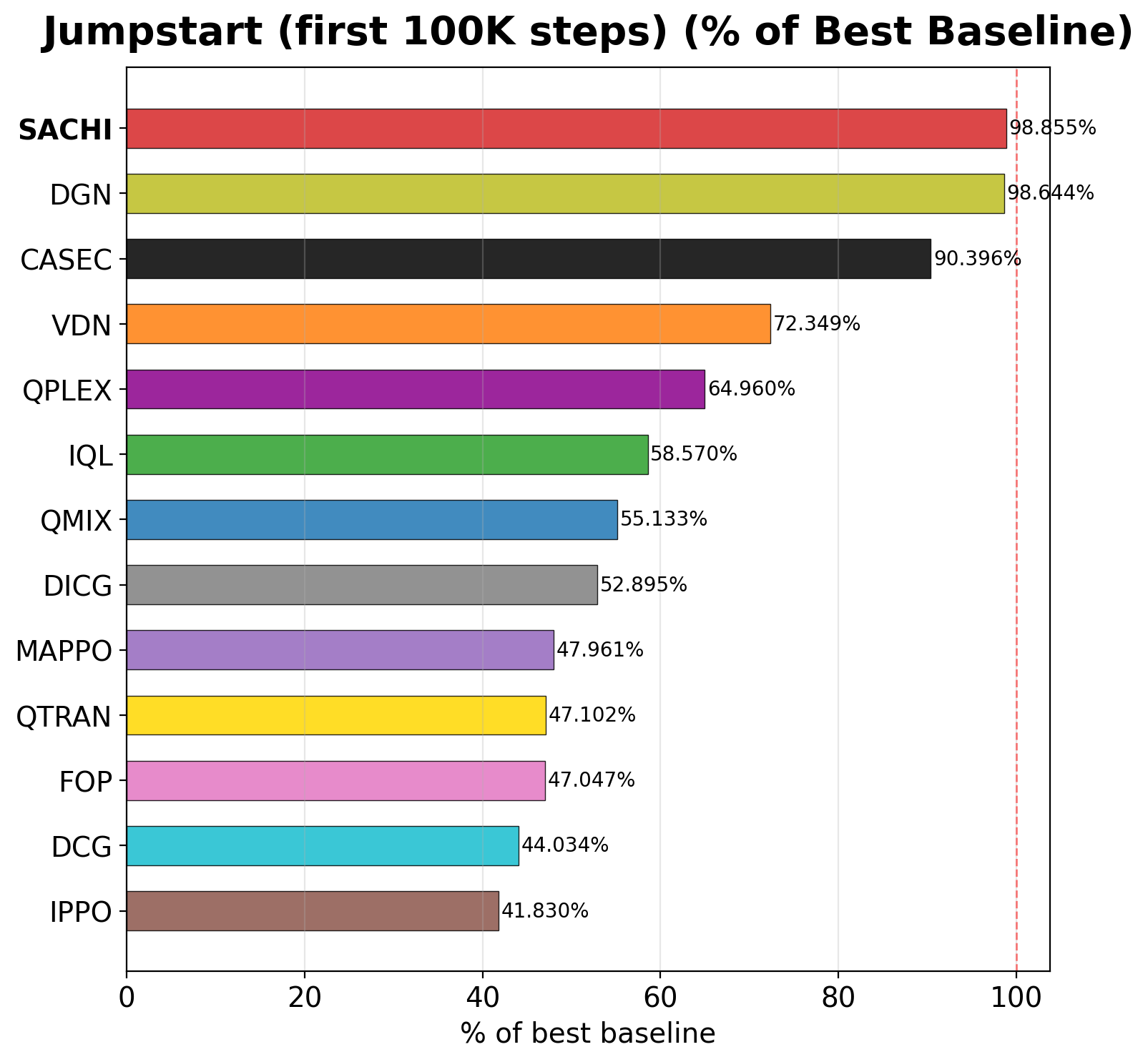}
    \caption{Jump-start performance over the first $100$K steps (percentage of best baseline's final score). \algo\ ranks first.}
    \label{fig:jumpstart}
% \vspace{-10pt}
\end{figure} 
A score of $0\%$ means the method matches the worst baseline's early performance; $100\%$ means it matches the best. The aggregate score is the mean of $\text{score}(m, e)$ across environments. Figure~\ref{fig:jumpstart} shows that \algo\ scores $98.9\%$, which is virtually indistinguishable from the best baseline's early performance on every environment. DGN ($98.6\%$) is the only method comparably close, while the remaining field drops off to CASEC ($90.4\%$), VDN ($72.3\%$), and QMIX ($41.8\%$). This implies that \algo's coordination module does not impose an early-training penalty: it begins producing useful inter-agent signals almost immediately, placing \algo\ among the fastest-coordinating methods from the start of training, not only at convergence.

\begin{figure}[!htp]
    \centering
    \includegraphics[width=0.8\linewidth]{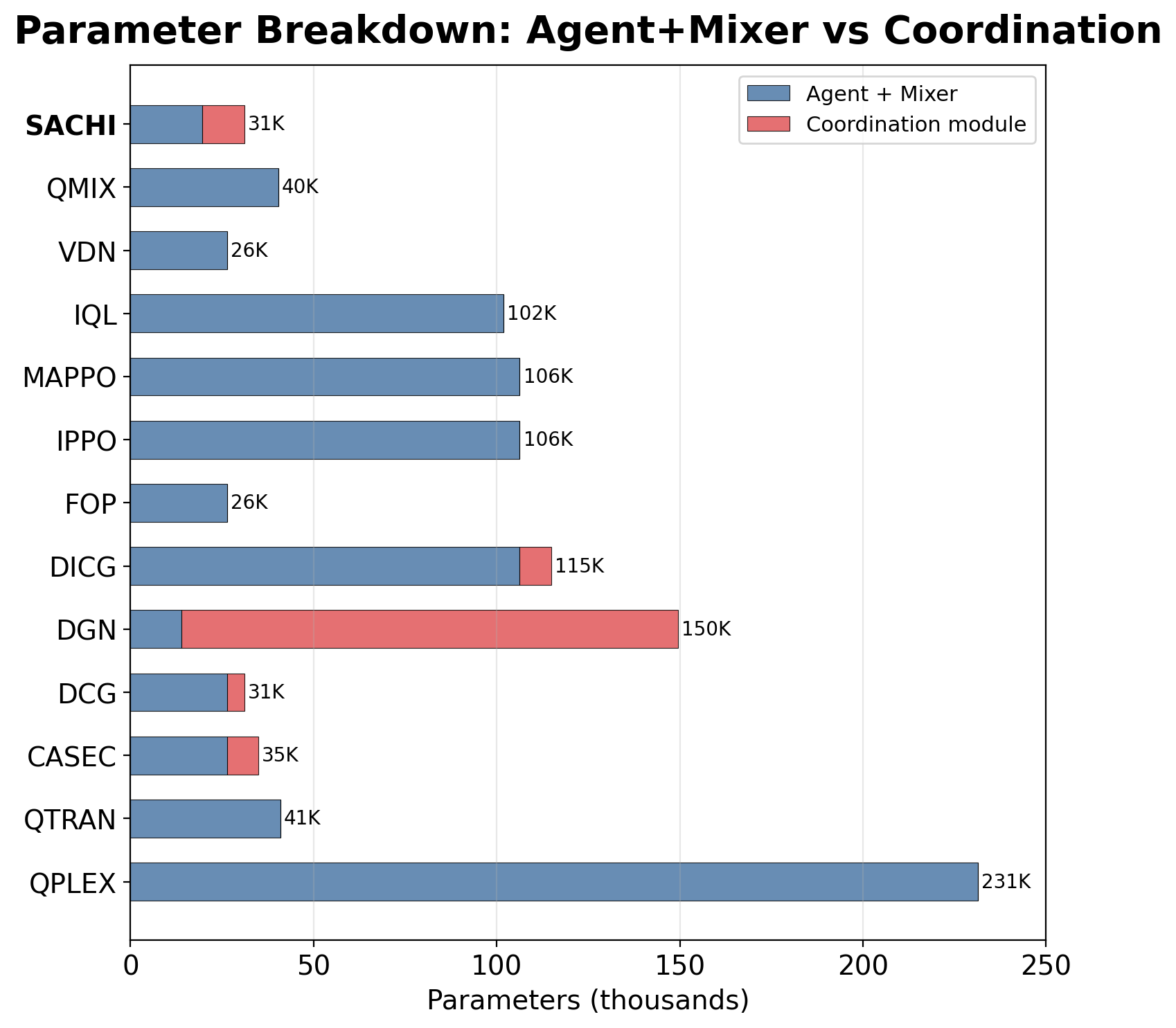}
    \caption{Parameter breakdown per method. Blue: agent + mixer. Red: coordination module. \algo\ has fewer total parameters than most baselines.}
    \label{fig:compute}
% \vspace{-12pt}
\end{figure} 

\begin{figure*}[!htp]
    \centering
    \includegraphics[width=0.85\linewidth]{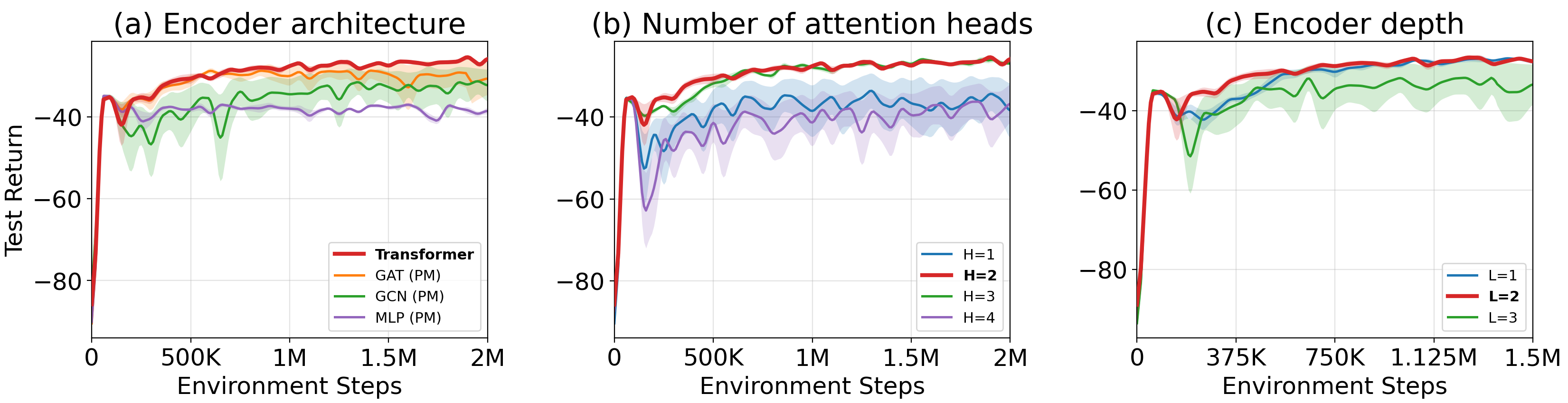}
    \caption{Ablation learning curves on \textsc{Reference}. (a)~Encoder architecture (parameter-matched). (b)~Number of attention heads. (c)~Layer depth. The default configuration is shown in red.}
    \label{fig:ablations}
% \vspace{-10pt}
\end{figure*} 

\paragraph{Parameter efficiency} Figure~\ref{fig:compute} decomposes each method's trainable parameter count into two components: (i)~shared infrastructure (agent network + mixer) and (ii)~the coordination module. For \algo, the latter includes the encoder, graph transformer layers, and aggregation network, contributing ${\sim}12$K parameters for a total of $31$K. Although \algo\ uses the same QMIX-style mixer, it is still smaller than QMIX ($40$K). The key difference lies in the agent network. QMIX and related value-decomposition methods rely on relatively heavy per-agent networks (e.g., GRUs, ${\sim}27$K parameters) to implicitly capture coordination effects and temporal dynamics within each agent's representation. In contrast, \algo\ makes coordination explicit via the graph transformer, allowing the per-agent network to be simplified to a lightweight feedforward model (${\sim}4$K). This architectural shift moves complexity from implicit, per-agent modeling to a shared coordination module, resulting in a lower overall parameter count. As a result, \algo\ remains smaller than QMIX ($40$K), QTRAN ($41$K), or MAPPO ($106$K), despite introducing an explicit coordination mechanism. QPLEX ($231$K) is the largest method in the comparison, more than $7{\times}$ the size of \algo, yet ranks last on both AUC and average rank. DGN, the strongest baseline on several tasks, requires $150$K parameters. This is nearly $5{\times}$ more than \algo, yet is consistently outperformed on the aggregate metrics. 

\subsection{Ablation Studies}
\label{sec:ablations}

The results above establish that \algo\ works. The ablations establish why it works. We vary three axes on \textsc{Reference} (the environment with the widest dynamic range across architectural variants) and report learning curves in Figure~\ref{fig:ablations}. 

\paragraph{Message-passing architecture (Panel a)} We replace Transformer with parameter-matched (PM) GAT, GCN, and MLP alternatives. The result is a clean monotone ordering: Transformer (${\approx}{-}26$) $>$ GAT (${\approx}{-}30$) $>$ GCN (${\approx}{-}35$) $>$ MLP (${\approx}{-}40$). This ordering is not accidental. It tracks exactly the degree of content-dependent message filtering each operator provides. The Transformer computes attention as a function of both sender and receiver states (Equation~\ref{eq:gtcg_attention}); GAT conditions on the sender but uses a shared receiver-independent scoring function; GCN applies a uniform transformation to all neighbors; and MLP ignores graph structure entirely. Moreover, parameter-matching ensures that these differences reflect inductive bias, i.e., how it processes inter-agent information, and not how many parameters it has. This is the central empirical result of the paper: it is not more parameters, more layers, or a fancier training procedure that drives \algo's advantage; it is content-dependent message passing. 

\paragraph{Attention heads (Panel b)} $K{=}2$ and $K{=}3$ yield stable convergence near $-26$, while $K{=}1$ and $K{=}4$ exhibit seed-dependent instability with standard-error bands spanning ${\sim}20$ points. The failure mode at $K{=}4$ is instructive: with four heads operating on $(d_{\text{in}}/4)$-dimensional subspaces, each head has a narrow view of the embedding, and the coordination signal becomes fragmented across heads that individually lack the capacity to capture the full inter-agent relationship. At $K{=}1$, a single head must represent all aspects of the coordination signal in one set of query-key projections, which succeeds on some seeds but not others. $K{=}2$ provides the minimum redundancy needed for robustness. 

\paragraph{Layer depth (Panel c)} $L{=}1$ and $L{=}2$ converge to similar final returns on \textsc{Reference}, where the coordination graph has diameter one and multi-hop propagation is unnecessary. $L{=}3$ degrades to ${\approx}{-}40$, consistent with oversmoothing: repeated neighborhood aggregation drives all agent representations toward a shared mean, destroying the individual information needed for decentralized action selection. For the team sizes in our evaluation ($n \leq 12$), $L{=}2$ provides a consistent trade-off: it covers each agent's $2$-hop neighborhood while remaining shallow enough to avoid representation collapse. This is consistent with findings in the broader GNN literature~\cite{shehzad2026graph,rusch2023survey,cai2020note,kipf2017semi}, where $2$--$3$ layers are typically sufficient even in graphs with thousands of nodes, and deeper stacks yield diminishing or negative returns due to oversmoothing.

\begin{table}[t]
\centering
\small
\setlength{\tabcolsep}{5pt}
\caption{Test return (mean $\pm$ stderr over 5 seeds) at training milestones on \textsc{reference}. Best per column in bold.}
\label{tab:ablation_more_samples}
\resizebox{\columnwidth}{!}{
\begin{tabular}{lcccc}
\toprule
Method & 2M & 5M & 8M & 10M \\
\midrule
\textbf{SACHI} & \textbf{-25.39 $\pm$ 0.35} & \underline{-23.94 $\pm$ 0.37} & \textbf{-22.63 $\pm$ 0.10} & \textbf{-22.17 $\pm$ 0.25} \\
\midrule
QMIX & -28.81 $\pm$ 0.39 & -26.29 $\pm$ 0.92 & -24.57 $\pm$ 0.73 & -23.84 $\pm$ 0.58 \\
VDN & -38.44 $\pm$ 0.39 & -35.70 $\pm$ 0.26 & -36.18 $\pm$ 0.14 & -36.41 $\pm$ 0.26 \\
IQL & -36.12 $\pm$ 0.33 & -35.01 $\pm$ 0.50 & -35.53 $\pm$ 0.14 & -36.07 $\pm$ 0.29 \\
MAPPO & -34.57 $\pm$ 0.07 & -33.59 $\pm$ 0.20 & -34.38 $\pm$ 0.28 & -35.76 $\pm$ 0.15 \\
IPPO & -34.69 $\pm$ 0.09 & -33.76 $\pm$ 0.20 & -34.10 $\pm$ 0.10 & -35.38 $\pm$ 0.09 \\
FOP & -35.38 $\pm$ 0.35 & -30.85 $\pm$ 2.35 & -28.75 $\pm$ 2.89 & -24.55 $\pm$ 1.36 \\
DICG & -34.97 $\pm$ 0.10 & -24.05 $\pm$ 0.19 & -23.37 $\pm$ 0.41 & -22.92 $\pm$ 0.31 \\
DGN & -41.95 $\pm$ 4.08 & -44.20 $\pm$ 7.24 & -43.10 $\pm$ 7.19 & -47.70 $\pm$ 10.24 \\
DCG & -27.34 $\pm$ 0.41 & \textbf{-23.33 $\pm$ 0.13} & -22.68 $\pm$ 0.17 & -22.50 $\pm$ 0.20 \\
CASEC & -50.71 $\pm$ 3.54 & -58.76 $\pm$ 3.51 & -61.98 $\pm$ 4.18 & -60.69 $\pm$ 4.24 \\
QTRAN & -39.33 $\pm$ 1.39 & -39.29 $\pm$ 0.61 & -39.52 $\pm$ 0.35 & -39.69 $\pm$ 0.55 \\
QPLEX & -65.72 $\pm$ 12.42 & -52.49 $\pm$ 6.36 & -64.28 $\pm$ 9.53 & -60.65 $\pm$ 7.09 \\
\bottomrule
\end{tabular}
} 
% \vspace{-16pt} 
% \vspace{-1em} 
\end{table} 

\paragraph{Extended training (Table~\ref{tab:ablation_more_samples})} We also extend training from $2$M to $10$M steps on \textsc{Reference} to examine whether the performance gap observed at $2$M reflects a lasting advantage of \algo\ or simply slower learning in competing methods. As training progresses, two patterns become clear. \algo\ improves consistently and retains the best performance at later checkpoints, indicating a strong asymptotic solution. At the same time, a subset of baselines, notably DICG and DCG, continue to improve with more samples and gradually narrow the gap. By convergence, the difference becomes small, suggesting that these methods can eventually reach similar performance levels given sufficient training. However, this convergence behavior highlights a more important distinction: learning speed. \algo\ attains high-quality solutions significantly earlier, while competing approaches require substantially more environment interactions to approach the same regime. In contrast, several other methods plateau or deteriorate with extended training, indicating that their limitations are not due to insufficient samples but to inherent representational constraints. 

\textbf{In summary,} across several cooperative environments, baselines, and statistical lenses, \algo\ emerges as the most consistently competitive method: average rank $1.90$, IQM at the baseline ceiling ($1.00$ with non-overlapping CIs), near-zero optimality gap ($0.01$), and stochastic dominance over every baseline at every performance threshold. It achieves this with fewer parameters ($31$K) than most baselines, the highest jump-start performance ($138\%$), and the highest learning-trajectory AUC ($1.01$). The ablations trace the source of this advantage to a single design choice: the parameter-matched ordering Transformer $>$ GAT $>$ GCN $>$ MLP shows that what drives \algo's performance is not model capacity but the degree of content-dependence in the message-passing operator, i.e., the ability of each agent to condition the information it extracts from a neighbor on both the neighbor's state and its own.

\section{Discussion}
\label{sec:discussion}

Section~\ref{sec:results} established what \algo\ achieves; this section examines why it works, what it costs, and where it applies. 

\paragraph{Why does \algo\ work?}
\label{subsec:why}
The consistent outperformance of \algo\ across diverse coordination challenges, from resource allocation (Disperse) to adversarial cooperation (Adversary) to directed communication (Speaker-Listener), points to three complementary explanations. First, the receiver-dependent attention mechanism (Eq ~\ref{eq:gtcg_attention}) provides each agent with a personalized view of its neighbors' states, filtered by its own current context. Ablations show a clear hierarchy (Transformer $>$ GAT $>$ GCN $>$ MLP) aligned with this capability. Second, the fully connected coordination graph with learned attention enables implicit structure learning: instead of fixing or explicitly pruning edges, the model learns which interactions matter, adapting to dynamic coordination patterns. 
% Third, residual connections preserve local observations while adding coordination signals, avoiding over-reliance on communication. Together, these lead to both effective coordination and faster early learning, as also reflected in strong jump-start performance. 
Third, the statistical evidence confirms that these architectural choices compound into a reliable advantage. The combination of aggregate dominance, per-environment consistency, and parameter efficiency rules out the most common alternative explanations: the gains are not driven by a single favorable task, by lucky trials, or by increased model capacity.

\paragraph{Computational Complexity}
\label{subsec:complexity}

The primary computational overhead of \algo\ relative to non-graph baselines arises from the graph transformer message-passing stage. For $n$ agents, $L$ layers, and embedding dimension $d$, each agent computes a query-key inner product with all $n - 1$ neighbors, requiring $\mathcal{O}(n^2 d)$ operations per layer. The full forward pass is therefore $\mathcal{O}(Ln^2 d)$ per timestep, compared to $\mathcal{O}(nd)$ for QMIX and $\mathcal{O}(Lnd)$ for GCN-based methods such as DICG. This quadratic scaling in $n$ is the standard cost--coordination trade-off in graph-based MARL. In practice, with $L = 2$, $n \leq 12$, and $d = d_{\text{in}}$ (typically small in our environments), the overhead is modest. However, the parameter analysis in Figure~\ref{fig:compute} reveals a perhaps counterintuitive fact: \algo's total parameter count ($31$K) is lower than QMIX ($40$K), QTRAN ($41$K), IQL ($102$K), DGN ($150$K), and QPLEX ($231$K). This is because the graph transformer's learnable parameters are fully shared across agents and independent of $n$, i.e., the total count is $\mathcal{O}(Ld^2)$, while \algo's feedforward agent network (${\sim}4$K) is substantially cheaper than the GRU used by many baselines (${\sim}27$K). The coordination module adds computational cost per forward pass but not model capacity in the sense that drives overfitting. For very large teams ($n > 50$), the quadratic attention cost may become prohibitive; sparsification of the coordination graph or hierarchical grouping strategies represent natural adaptations.

\paragraph{Scalability}
\label{subsec:scalability}

Scalability in cooperative multi-agent systems concerns maintaining effective coordination as interaction size and complexity grow. 
% We analyze scalability along three axes in our evaluation suite: . 
The key design principle of \algo's coordination mechanism is disentangled from common scalability axes such as team size, action space, and observation structure. It operates over a permutation-equivariant graph with shared parameters, while task-specific variability is handled at the input and output interfaces. Across environments ranging from $n = 2$ to $12$, \algo\ uses the same architecture without structural modification, with parameter sharing inducing permutation equivariance, i.e., for any permutation $\pi$, $\tilde{\mathbf{h}}_{\pi(i)} = f(\mathbf{h}_{\pi(1)}, \ldots, \mathbf{h}_{\pi(n)})_{\pi(i)}$. 
% As in QMIX, whose mixing network depends on the global state rather than agent identity, this supports scaling across team sizes without additional tuning. 
Furthermore, the evaluation suite spans movement, communication, and hybrid tasks with varying observation dimensionality. The encoder $\phi_{\text{enc}}$ maps observations to a fixed embedding, decoupling message passing from input size, while the policy head $\psi$ localizes dependence on $|\mathcal{A}_i|$, making the coordination module invariant to both action and observation dimensions. Despite variation in observation structure (e.g., role asymmetry, adversarial signals, symmetry-breaking), the query-key attention adapts implicitly as $\alpha_{ij} \propto \exp((\mathbf{W}_3 \mathbf{h}_i)^\top (\mathbf{W}_4 \mathbf{h}_j)/\sqrt{d_{\text{in}}})$, evaluating the relevance of agent $j$'s signal from agent $i$'s perspective without task-specific assumptions. 

\paragraph{When to apply \algo? And when not to?}
\label{subsec:applicability} 
\algo\ is most effective in cooperative settings where useful information is distributed across agents and must be selectively integrated for decision-making. It is well-suited for tasks with partial observability, dynamic or context-dependent coordination structure, and heterogeneous agent roles, such as communication tasks, spatial coordination, and mixed cooperative-adversarial environments. In these settings, its receiver-dependent attention and adaptive coordination graph enable agents to extract the right information from the right peers at the right time, leading to both faster learning and strong final performance. \algo\ can be considered less beneficial in settings where coordination is minimal or trivial, such as fully observable environments or tasks that decompose cleanly into independent subproblems. In such cases, simpler methods (e.g., independent learners or value decomposition) may suffice with lower computational overhead. Additionally, in very large-scale systems with strict communication or latency constraints, the intricate attention mechanism may introduce scalability challenges unless appropriately pruned or approximated.

\section{Conclusion and Future Work}
\label{sec:conclusion}

In this paper, we proposed \algo\ for cooperative MARL to address action coordination as a policy-level problem of structured information
integration. By enriching each agent's representation with coordination-relevant signals
from teammates through graph transformer convolutions over an inter-agent coordination
graph, \algo\ enables decentralized agents to jointly produce coherent action profiles. The key architectural insight is that effective
coordination requires not the collection of all available information, but its intelligent,
receiver-sensitive distillation: the graph transformer's content-dependent attention allows
each agent to selectively accumulate precisely what it needs from its neighbors, given its
own current state. Empirical evaluation across five cooperative tasks, spanning spatial,
communicative and adversarial coordination challenges, demonstrates consistent improvements
over twelve baselines covering the full range of coordination methodologies, and ablations confirm that the gains are attributable to the inductive bias of the graph
transformer rather than increased model capacity. We hope this work encourages further
treatment of action coordination as a first-class problem in cooperative MARL, and
anticipate that the structured information integration perspective developed here will
generalize to richer settings involving larger agent teams, heterogeneous roles, and
partially known coordination topologies.

\section*{Acknowledgment}

This work is supported by DEVCOM Army Research Office; Grant: W911NF2420194; and U.S. National Science Foundation (NSF); Grant: OAC-2411446. Distribution Statement A: Approved for public release. Distribution is unlimited. 

% References
\bibliography{main} % bibliography data in report.bib
\bibliographystyle{IEEEtran} % makes bibtex use spiebib.bst

% \clearpage  
% \section*{Appendix}
% \onecolumn
% \label{sec:Appendix}
% \input{sections/10_appendix} 

\end{document}